\newif\ifpdf

\ifx\pdfoutput\undefined
   \pdffalse 
\else
   \pdfoutput=1 
   \pdftrue
\fi

\documentclass[12pt,a4paper]{article}
\ifpdf
   \usepackage[pdftex,xdvi]{graphicx}
   \usepackage[pdftex]{color}
\else
   \usepackage[dvips,xdvi]{graphicx}
   \usepackage{color}
\fi

\usepackage[isolatin]{inputenc}

\begin{document}

\noindent{\Large\bf Neural Architectures for Robot Intelligence}

\bigskip
\renewcommand{\thefootnote}{\fnsymbol{footnote}}
\noindent{\bf\large H. Ritter\footnote[1]{to whom correspondence should be addressed (helge@techfak.uni-bielefeld.de)}, J.J. Steil, C. Nölker, F. R\"othling, P. McGuire}
 
\renewcommand{\thefootnote}{\fnsymbol{arabic}}
\medskip

\noindent{Neuroinformatics Group, Faculty of Technology, Bielefeld University,} 

\noindent{POB 100131, 33501 Bielefeld, Germany}

\bigskip
\noindent{\bf Abstract:} We argue that the direct experimental approaches to elucidate the architecture
of higher brains may benefit from insights gained from exploring the possibilities
and limits of artificial control architectures for robot systems. We present
some of our recent work that has been motivated by that view and that is 
centered around the study of various aspects of hand actions since these
are intimately linked with many higher cognitive abilities. As examples,
we report on the development of a modular system for the recognition of
continuous hand postures based on neural nets, the use of vision and tactile
sensing for guiding prehensile movements of a multifingered hand, and the 
recognition and use of hand gestures for robot teaching.

Regarding the issue of learning, we propose to view real-world learning
from the perspective of data mining and to focus more strongly on the 
imitation of observed actions instead of purely reinforcement-based
exploration. As a concrete example of such an effort we report
on the status of an ongoing project in our lab in which a robot
equipped with an attention system with a neurally inspired
architecture is taught actions by using hand gestures in 
conjunction with speech commands. We point out some of the lessons
learnt from this system, and discuss how systems of this kind
can contribute to the study of issues at the junction between
natural and artificial cognitive systems.




\vfill
\newpage


One of the most remarkable feats of nature is the evolution of the
information processing architectures of brains. Despite the use of
components that are -- as compared to transistor devices -- slow,
of low accuracy and of high "manufacturing tolerances", evolution has
found architectures that operate on highly complex perception and
control tasks in real time, by far outperforming our most sophisticated
technical solutions not only in speed, but also in robustness and
adaptability.

If the main architectural feature that leads to such properties
were as simple as "massively parallel processing", 
the current capabilities of artificial systems should look much more
advanced than they actually are. While the raw processing power of
modern microprocessors -- if used in larger numbers -- begin to 
become comparable with that of sizeable portions of brain tissue, 
we believe that we still lack most of the concepts that are required to shape this
processing power into brain-like capabilities.

\section{The challenge of brain architecture\label{Sec:challenge}}

Much of brain research in the past and in the present has
-- often dictated by experimental constraints -- emphasized a bottom
up approach in which the properties of individual neurons or even
synapses are in the center. Recent years have seen the advent of
methods that can significantly help to complement this with a more
top-down approach.

We may argue that the work of recent years and decades has
now accumulated to the point where we may form quite reasonable guesses
about the computational contribution of quite a substantial number
of brain areas. While a more detailed elucidation of the exact nature
of these computations may still need a long time of further research,
we may nevertheless have the hope that many of these details are not decisive for
the operation of the overall system.

\begin{figure}[h!]\centering
\includegraphics[height=0.6\textwidth]{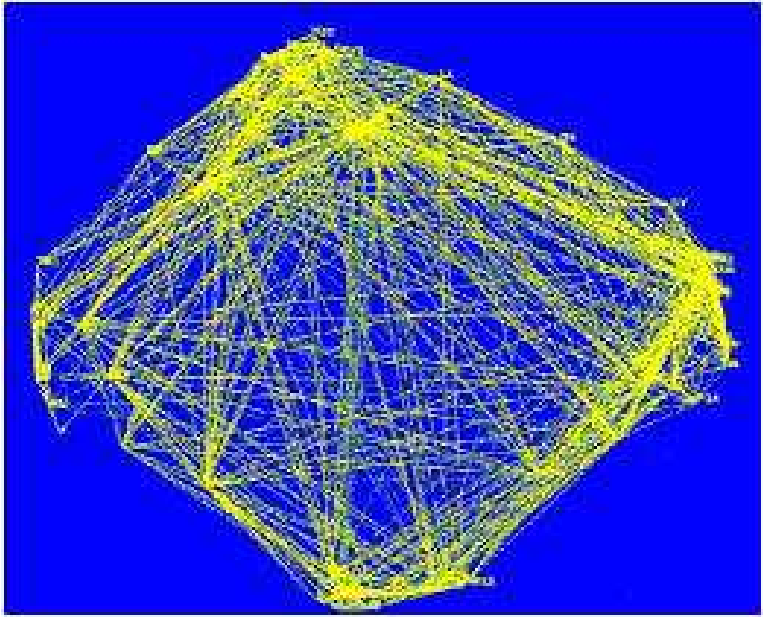}
\caption{Multidimensional scaling plot of connectivity between functional 
areas in the cat brain (from \cite{Young93}): functional areas are indicated by labelled nodes,
major connectivity bundles by lines. The technique of multidimensional 
scaling has been used to map brain centers to node positions in such
a way that spatial proximity between nodes reflects a high degree
of mutual interconnectivity.}\label{Fig:cat}
\end{figure}

In support of this view are the results of recent studies of brain
organization that have revealed a rather tight interconnectivity of the
majority of brain centers \cite{NI00,Young94,Young93}, 
making it likely that it is the structure
at this level that shapes system properties much more than the details
in the modules themselves. Fig. \ref{Fig:cat} from \cite{Young93}
depicts a view of the intermodule-connectivity in cat brain.
Nodes indicate major processing centers and each line represents
a major interconnection bundle. The technique of multidimensional
scaling has been used to depict processing centers at positions in
such a way that their spatial distance reflect their degree of
their interconnectedness. Diagrams like this can summarize huge amounts of
anatomical data and give us a ''bird's eye view'' on brain
architecture.

In fact, from technical control systems it is well-known that heavy feedback
connections can almost entirely determine the dynamic system behavior,
effectively ``disguising'' almost all dynamical details (and imperfections!)
that may be exhibited if the modules would operate in open loop mode.

Therefore, even rather crude approximations to the different brain
module functions may give us a chance to gain significant insights 
into the functioning of the overall system, provided we get the 
architectural level right. If this view is correct, a significant
part of the challenge will be the exploration of possible architectural
patterns that could organize the computational contributions of a
collection of sensor-, motor- and memory modules into a coherent
sensori-motor-processing activity.

While controlled experimental alterations of actual brain architectures
are a very difficult -- and at least in higher animals ethically highly
objectionable -- task, the domain of robotics offers much wider
experimental possibilities: current technology has matured to the point where we
can (admittedly only very coarsely in many cases) approximate a reasonable 
spectrum of isolated perceptual, memory, and motor capabilities, allowing
us now to explore architectures for the integration of these functions
into artificial cognitive systems. 

Behind such an approach is the expectation that the computational 
architecture of brains is to a large extent shaped by the computational 
demands of the tasks that they solve (certainly, there will be other
significant factors, such as the available kind of ``hardware''; however,
at least in the technical domain the experience is that software can
compensate for a wide range of different hardware characteristics).

If this is indeed the case, it will be important to explore computational
architectures primarily for those tasks that are likely to be typical 
for what brains have to solve. From an evolutionary perspective, one
likely candidate is navigation, which is one of the most basic behaviors
that contributes to the capabilities of an animal. However, the example
of insect navigation also shows that the evolution of cognition 
apparently has been driven
by demands that are additional to those that gave rise to 
the ability of navigation alone.

\section{Hands and Actions\label{Sec:hands}}

It has been argued that the development of  cognition is closely 
linked with the capability to purposively act on one's environment and to cause
changes to it \cite{TheSmi94}. Therefore, we may expect that the need to control 
sophisticated manipulators, particularly in the form of arms and hands 
(for many animals, also the mouth will play the role of an important
``mani''pulator) will be a major driving force and shaping factor
for the cognitive processing architecture of a brain.
In fact, a closer look reveals that in particular the control of
hands in the human is connected with a large number of highly demanding
and in many ways generic information processing tasks \cite{HandAndBrain96}
whose coordination
already forms a major base for intelligent behavior.

\begin{figure}[ht!]\centering
\includegraphics[height=\textwidth,angle=-90]{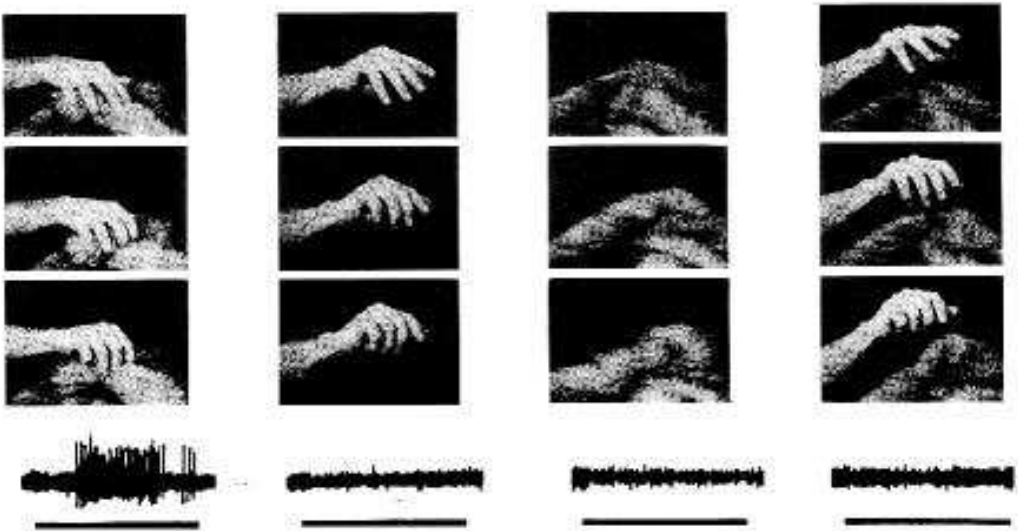}
\caption{Selective neuron response to visually perceived hand actions (from
  \cite{Perret89}). 
  The bottom row shows the associated activity trace of a neuron (the same for all trials) in
  the STS (superior temporal sulcus) of a macaque when the animal observes the
  action sequence (1 s duration, indicated by calibration bar at the bottom) 
  depicted in the corresponding column above.
  Apparently, the recorded neuron responds very selectively to the interaction
  between hand and fur (leftmost column) and is silent in the remaining, visually
  very similar trials.
}\label{Fig:hand+fur}
\end{figure}

Regarding perception, a first major issue is the visual recognition of hands
and hand actions. Since hands are a major focus of action, we -- but also
other higher primates -- are very good at the visual recognition of hands,
their motions as well as their highly variable postures and -- at an even
higher level -- their actions. Most interestingly, such recognition tasks
were found to be correlated with the activity of neurons (located in the 
superior temporal sulcus) that show rather selective responses to 
visually perceived hand actions \cite{Perret89}, see Fig. \ref{Fig:hand+fur}.
Therefore, the task of recognition of hands and
their actions is a very suitable starting place to investigate
a minimal set of demands on the architecture of a visual system, and
in Sec. \ref{Sec:grefit} we will report on ongoing work in our lab towards the goal
of visually based recognition of hand actions.

Touch is another modality closely linked with our hands. The haptic perception
of objects is highly developed in humans and it is known that some of the
initial processing steps in somatosensory cortex (edge filtering etc.) 
resembles visual processing in V1 \cite{JHT95}. 
However, there are also significant
differences: haptic perception is usually closely coupled with finger
movements during which -- unlike the retinal surface in our eyes -- the skin of
our fingers undergoes complex changes in its three-dimensional arrangement
in space and relative to the object surfaces. From a computational point
of view, this poses a formidable challenge which no doubt can only be
addressed at the level of an architecture that is suitable to coordinate
the operation of tactile sensors and active finger control. 
In recent years, we have developed finger tip force sensors 
\cite{JockuschWalterRitter1997-ATS} that
provide at least some rudimentary touch force information for the
finger tips of a three fingered hand and in Sec. \ref{Sec:haptics} we will briefly
report how this research can help to approach issues such as
how to control finger synergies for grasping and fusion of touch with
other sensory modalities, such as vision.

Finally, there is the issue of carrying out hand actions themselves.
This increases the demands on the underlying architecture even further:
besides the integration of perception and action we have as additional
elements the need for some state and context memory as well as planning.
The neural structures involved in these stages appear to 
be farther removed from direct sensory input or motor output and include
areas such as the 
parietal cortex, the premotor and the supplementory motor area. 
To carry out their task, these systems must operate on highly preprocessed
representations that reflect important invariants at the task level,
such as holding an object, aligning it with another one or controlling
a constrained movement, such as screwing a nut onto a screw.

At the same time, the operation of these structures exhibits an
enormous flexibility and it is only the ability of learning that can
make possible and account for the enormous range of human manual activities.
Therefore, a major challenge at this architectural level is to understand
how neural adaptivity at the lower system levels can manifest itself
as a progressive shaping of the interactions among neural modules in the 
large. In other words, we have to understand what principles are required
to realize a ``learning architecture'' in which a substantial number
of coarsely adapted skills can become coordinated in different ways,
such as to allow goal-directed sequences of manual actions. 

Research on unsupervised reinforcement learning algorithms has tried to
address this issue, however, with limited success. While there have been
some remarkable demonstrations of unsupervised skill learning on simple 
tasks, the learning of more complex tasks usually leads to exponentially growing
learning times and quickly becomes unrealistic even for tasks of moderate
complexity \cite{kaelbling96reinforcement}. 
Additionally, the success of these demonstrations
usually relies on often rather ingeniously encoded input data. Typical
real world learning situations differ markedly from this: they are usually
characterized by extremely high-dimensional sensory input, such as vision
or touch, and it is part of the challenge for the learning system to find
out which regularities in the input can guide successful execution
of the intended task.

The task of detecting such regularities shares many issues with the 
still rather young field of datamining: in datamining, the goal is to
detect patterns and regularities in often huge amounts of data in order
to support decisions, make predictions or improve the control of some
industrial process \cite{fayyad96,adrians96}. 
In a way, modern datamining systems can be seen as 
the analogous endeavor carried through by nature when evolving brains:
to endow a company, similar to an organism, with sensors and perceptual
capabilities to exploit useful regularities in its surround
in order to improve its fitness. While brains have been evolved for
the processing of sensory signals we are largely familiar with,
datamining techniques attempt to imitate similar capabilities -- although
at a currently much lower level -- for data from artificial, man-made
domains, such as economic systems, medical databases, industrial
processes, or the internet.

Therefore, it is not too surprising that many methods in the field of datamining
are closely related to the mathematics that underlie many models of brain operation. 
To exploit these analogies may be fruitful for both sides: brain modeling
may benefit from the rapid progress of current datamining methods, and
insight into brain mechanisms may inspire new datamining techniques.

However, it seems doubtful that this alone is sufficient to realize
a high-level learning architecture. Even the most sophisticated machine
learning methods are unlikely to remove the ``curse of dimensionality''
that limits the possibilities of unsupervised learning in high-dimensional
spaces. Therefore, it appears to us that the only promising way towards
high-level learning is to detect the required high-level structures
in the input data themselves, instead of trying to completely re-construct them
by search in a high-dimensional search space.

 
An attractive approach along these lines -- and again intimately 
connected with the observation and interpretation of hand actions --
is offered by the paradigm of {\em imitation learning} 
\cite{Kuniyoshi94,Bakker96}.
Watching an action sequence offers a learner the chance to 
strongly reduce the search space he or she is confronted with and
to focus exploration to a more or less narrow neighborhood of a successful
example of the to-be-learned skill.  To make this feasible,
however, requires to identify essential elements of the action
in the sensory stream and to remap them into the context of the
observer. The first step requires a highly developed vision capability
(possibly supported by acoustic perception),
with the ability to detect relations in image sequences. The second step
requires a matching of the observed situation to one's own context; this
goes well beyond a simple geometric transformation, since the 
involved objects and relationships between them usually will only
be analogous but not identical to the situation in which the learner
wants to operate. Therefore, the approach of imitation learning
can only be implemented on top of a rather highly developed base
of sensory preprocessing that ``opens the door'' to detect useful
regularities in a complex world. It may turn out that this is the price to
pay for making learning feasible and that approaches that try to
solve ``the'' learning problem by ``more intelligent'' search strategies alone
are doomed to fail when it comes to the complexity of real world tasks, 
since they address the much harder problem of ``invention'' as compared
to ``imitation''.

\section{Neural net-based hand posture recognition\label{Sec:grefit}}

\begin{figure}[ht!]\centering
\includegraphics[height=4cm]{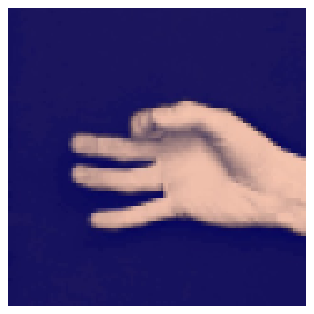}\hspace{0.15\textwidth}
\includegraphics[height=4cm]{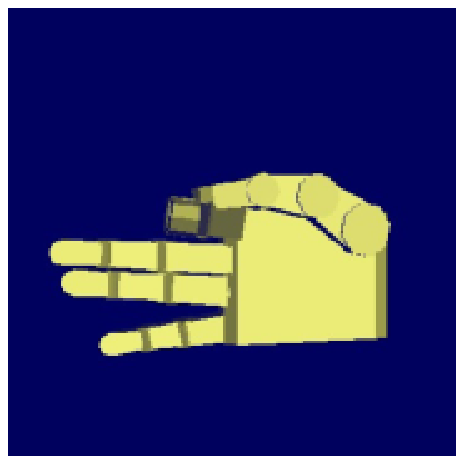}
\caption{Task of artificial neural recognition system: extract from
         2D monocular video image of a human hand (left) its 3D shape in order
         to permit tracking of the ''seen'' hand posture by 
         a computer rendered articulated hand model (right).
        }\label{Fig:ColGest}
\end{figure}

Recognition of 3D hand postures constitutes one of the complex
tasks that are routinely and effortlessly carried out by our brains.
In contrast to most everyday objects, our hands are deformable
objects, characterized by about 20 continuous degrees of freedom
(33 muscles actuating 20 joints).
The associated, very large configuration space makes the recognition
of continuous hand postures a much harder problem as, e.g., classification
of only a set of discrete hand configurations.
As a result, robust and accurate recognition of hand postures requires
to address many issues of more general vision systems. Still,
this task is sufficiently circumscribed to make the implementation
of a complete recognition architecture feasible. In the following,
we will describe the architecture of a hierarchical recognition system 
that employs several artificial neural nets in order to 
extract from monocular video images of a human hand an explicit
representation of the three-dimensional hand posture so that a
computer-rendered articulated model of the hand can follow the
posture of the viewed human hand (Fig.\ref{Fig:ColGest}).

In line with our previous remarks, the described recognition system  
(for more details, see \cite{NoeRit02}) 
does not attempt to follow any biological detail. Instead,
it is meant to explore a particular processing architecture whose 
modules coarsely mimic some of the principles that are thought to underlie
processing in the visual system. This allows then to study the significance
of these principles for the processing properties at the system level,
such as the contribution of multiple processing streams to achieve
better robustness, or the use of a hierarchical coarse-to-fine strategy
to focus on important parts of the image.

\begin{figure}[t]\centering
\includegraphics[width=0.8\textwidth]{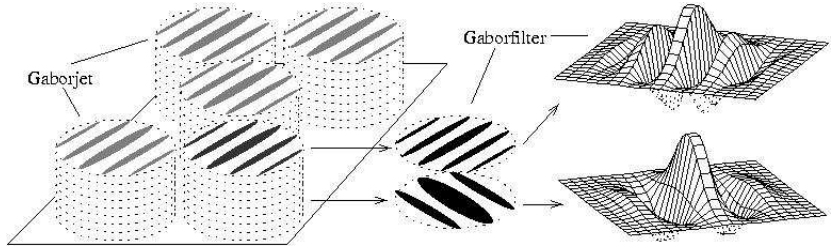}
\caption{Schematic arrangement of gabor jets in image plane (left) and
spatial profile of two different gabor functions (right).}\label{Fig:gaborjet}
\end{figure}


As a first step, the pixel image is transformed into a lower-dimensional
feature vector of ``neural activities''. We use a discrete grid of image
positions and position at each such location a number of ``receptive
fields'' of formal neurons that are chosen as Gabor functions
with different resolution and orientation (see Fig. \ref{Fig:gaborjet}).

The choice of these functions is motivated by the observation of Gabor-like
response characteristics in visual neurons (\cite{Dau80}) and also
by favorable mathematical properties of Gabor functions to capture important
local image information.

The resulting activity pattern provides an initial, ``holistic''
input representation of the image. While we have studied the extraction of hand posture
directly from such holistic activity patterns by means of neural learning
algorithms \cite{MeyRit92b}, we found that such a simple ``single-stage architecture''
is rather limited in the achievable accuracy of hand posture identification.

\begin{figure}[b!]\centering
\includegraphics[width=0.45\textwidth]{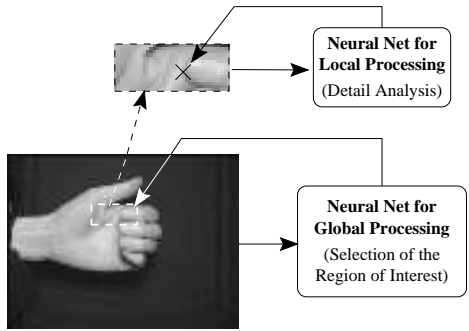}
\caption{Two-level processing hierarchy for determining finger tip
location: for each finger, a lower level network determines a ``focus region''
in which an upper level network attempts to determine the finger tip location. 
}\label{Fig:grefit_levels}\label{Fig:GlobalLocal}
\end{figure}


Experience with that approach has led us to the introduction of two 
additional ingredients into our vision architecture: $(i)$ instead
of a direct computation of the final hand posture, we attempt to first
extract from the holistic input representation a set of meaningful and 
stable object features which we chose
in the present task to be the 2D finger tip locations in the image.
$(ii)$ the solution of this subtask is not attempted in a single step;
instead, we use a processing hierarchy in which a neural network operating
on the holistic input representation first computes a coarse estimate of
the centers of up to five image subregions (one per finger) where the finger tips should 
be located. Subsequently, and on the second level of the processing hierarchy,
we provide for each finger subregion an extra network, applying an 
analogous processing as on the first
stage, but entirely focused on the selected finger subregion that was
identified by the first, ``global'' network (i.e., any visual 
input outside this region is clipped), and using Gabor functions at a
correspondingly reduced length scale (Fig. \ref{Fig:GlobalLocal}).  

Each network is itself modular and combines characteristics of a
fully localized representation (feature selective cells sharply
tuned to different pattern prototypes: ``grandmother cells'') and
a fully distributed representation (broadly tuned feature selective
cells such that only the combination of many activities is meaningful).

\begin{figure}[ht!]\centering
\includegraphics[width=0.4\textwidth]{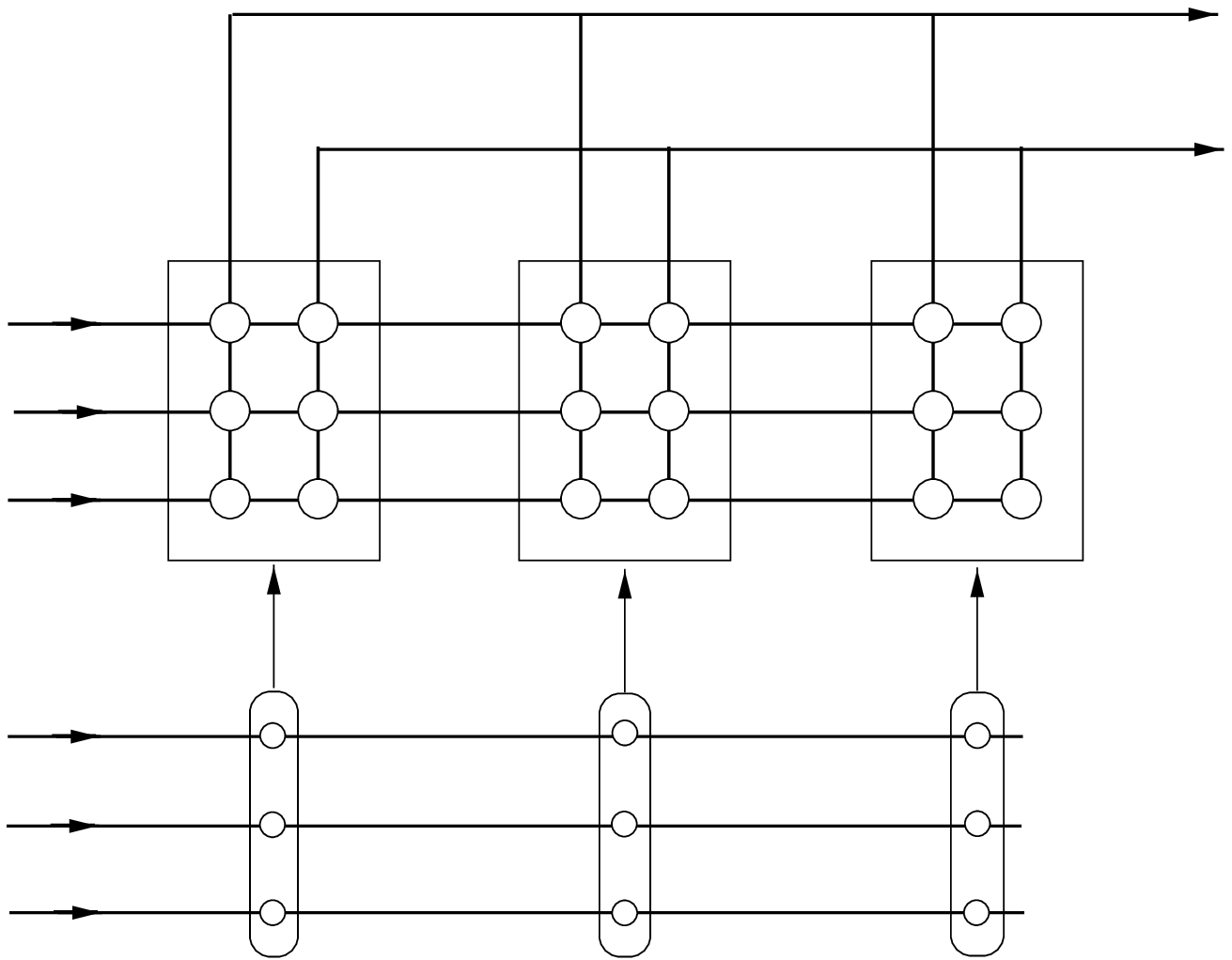}\hspace{0.05\textwidth}
\includegraphics[width=0.4\textwidth]{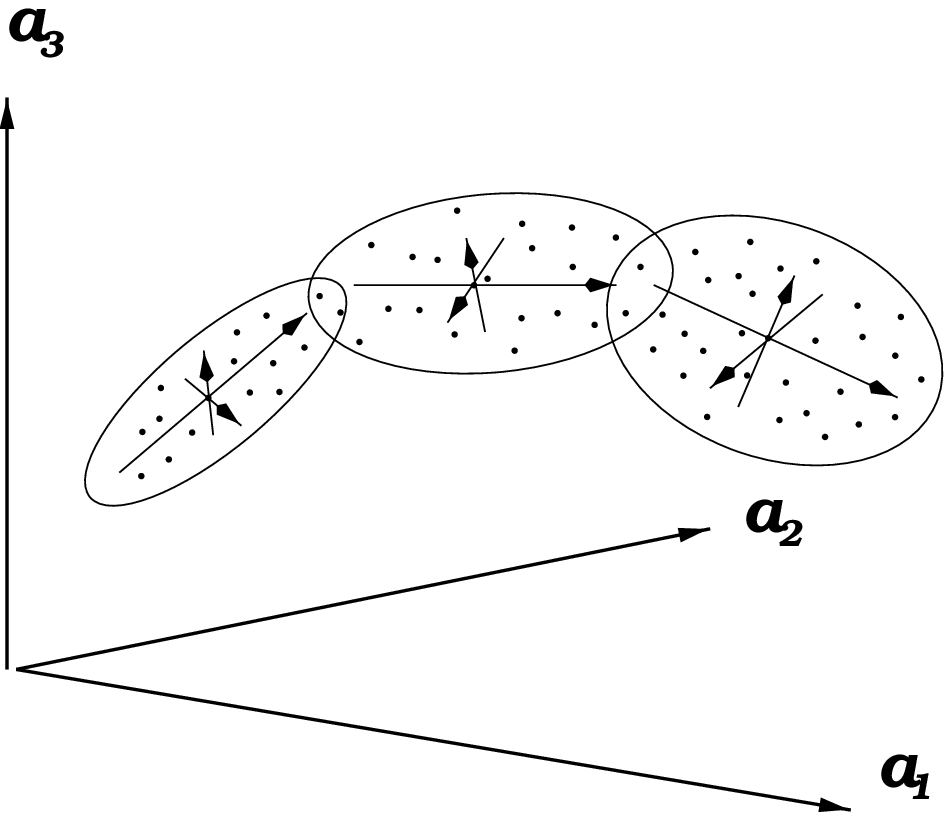}
\caption{{\em Left:} Local Linear Map (LLM) networks for supervised learning. Each
locally valid linear map is implemented as a linear perceptron (upper layer) that
is activated only for a subregion of the input space. The subregions are defined
by a layer (bottom) of competitive ``gating neurons''. {\em Right:} 
  PCA analysers for unsupervised classification. Each PCA analyser is valid
only for a subregion of the input space and orients its axes along the 
directions of maximal variance of the data distribution in its subregion.
}\label{Fig:LLMnet}
\end{figure}


\begin{figure}[t!]\centering
\includegraphics[width=0.8\textwidth]{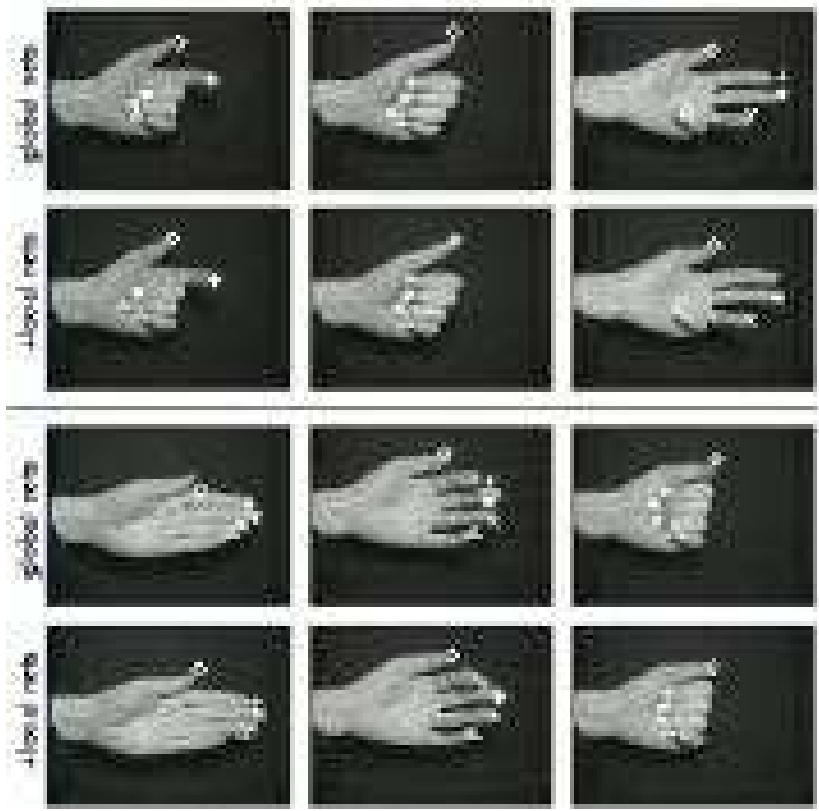}
\caption{Accuracy of finger tip detection without use of hierarchy (global nets only:
rows 1 and 3) and improvement through two-level hierarchy (+local nets: rows
2 and 4).}\label{Fig:grefit_mono}\label{Fig:Performance}
\end{figure}

The underlying mathematical principle of these networks is the
representation of an input-output mapping through a combination of
local linear maps in the case of supervised learning (Fig. \ref{Fig:LLMnet} (left)), 
or the representation
of a stimulus density by a collection of local principal component
analyzers in the case of unsupervised learning (Fig. \ref{Fig:LLMnet} (right)).

Both situations share the idea of using a layer of ``competitive''
neurons (the ``grandmother'' cells) to tessellate the feature space 
into a number of smaller and thus more manageable subregions and
to employ within each subregion a locally linear representation
(local perceptron-type or PCA-type mapping, employing a local
subnetwork of formal neurons). This type of approach has become
very popular during recent years and we have considered suitable
training algorithms in more detail elsewhere \cite{MeinickeRitter2001-RBC}. 
For the present situation,
the required training data consist of a sufficient number (several
hundred) of monochrome hand images for which the finger tip $(x,y)$-locations
must be explicitly provided (e.g., labeled by a human).
The first (``global'') network is directly trained with these data;
subsequently, each network of the second stage is then trained 
with the image subregions that are identified by the trained 
global network of the first stage (again using the associated
$(x,y)$-pairs as target output values). Fig. \ref{Fig:Performance}
shows for six typical hand postures the extracted finger tip locations 
obtained from the global network alone (rows 1 and 3) and the 
improved locations after including the corrections by the local 
networks (rows 2 and 4).

\begin{figure}[t]\centering
\includegraphics[width=0.6\textwidth]{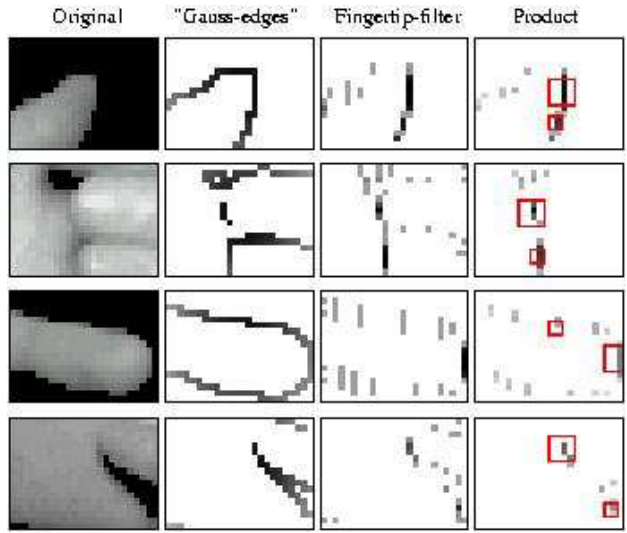}
\caption{Pixelwise multiplication of different filter results to
  integrate different processing streams (for details see text).}\label{Fig:knowledge}
\end{figure}


\begin{figure}[b!]\centering
\includegraphics[width=0.5\textwidth]{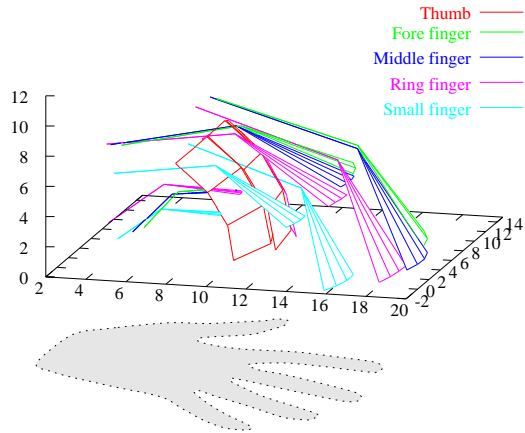}
\caption{PSOM-based mapping from fingertip positions to hand postures.
The bottom rectangle coincides with the viewing area enclosing 
hand shapes, the vertical axis denotes image depth. Each fan-like
grid structure belongs to one finger tip and indicates a set of 4x4 
finger tip positions traversed when independently varying the two
joint parameters (per finger) over 4 discrete values. The resulting
point set is used as training data for the PSOM network that achieves
a smooth interpolation through the depicted fan-like structures,
thereby yielding a smooth relationship between (observable) finger tip
positions in the viewing (=bottom) plane and (unobservable) finger tip
depth (vertical axis) as well as associated joint parameters (
interpolated fan-grid coordinates).
}\label{Fig:gridview3col}
\end{figure}

To further increase the accuracy of the obtained finger tip locations
we employ another architectural principle known to exist in the brain:
the fusion of multiple processing channels \cite{EssenDeYoe95}. For the present task, we
fuse three different processing ``streams'': the first stream is
obtained by a Gaussian
activity profile that is centered at the finger tip location determined
by the 2nd stage detection network for that finger. This activity profile
represents a confidence measure for the presence of the corresponding
finger tip that achieves its maximum at the estimated position, but decays
smoothly as one moves away from that point.

The second processing stream is motivated by the observation that 
(for the considered hand orientation) finger tips tend to be located
at intensity edges. Thus, we compute the second processing
stream as the edge image of the input distribution. Fig. \ref{Fig:knowledge}, 2nd column
shows the product of the first two processing streams, which is an
intensity modulated edge with maximum modulation at the point of
closest approach to the finger tip position as estimated from the network.

We convolve the input image with a simple ``finger tip filter'' that consists of
a 5x5 pixel template of a typical ``finger tip edge'', i.e., a short
vertical line with a two-sided (to take care of right and
left pointing directions) rounding at both ends (i.e., the template resembles
a ''{\tt )(}''-figure, with the middle parts of the brackets coalesced into a common
vertical edge) to obtain 
a 3rd processing stream (Fig. \ref{Fig:knowledge}, 3rd column).

All three processing streams are normalized to yield images with confidence
values in between 0 and 1 and are then pixel-wise multiplied to obtain the final
result (Fig. \ref{Fig:knowledge}, rightmost column). The two highest values in the 
result
image are then used as the most probable position candidate and an alternative
position (large and small squares in Fig. \ref{Fig:knowledge}).


In the final step, the obtained 2D features (finger tip locations)
are used to identify the 3D hand posture. There is a large debate to
what extent our vision system actually achieves a 3D reconstruction of perceived
objects (see, e.g., \cite{Peters00} for a survey). 
While there are many arguments that the available intensity
pattern of most real world images is subject to too many degradations
to allow a full and general 3D shape reconstruction by local algorithms
(``shape from X'' approaches), it seems that the visual system can very
efficiently use the available 2D views as an ``index'' into a 3D shape memory
that then provides access to 3D object properties. While little is
known about the precise nature of such a shape memory, on a more general
level it can be viewed as providing a set of shape models that 
provide sufficient constraints to make the available 2D information
interpretable in 3D.

Within our architecture, the available 2D information consists of
five coordinate pairs $(x_i,y_i)$, $i=1\ldots 5$, one for each finger.
Obviously, this is insufficient for a 3D-reconstruction even of the
finger tip positions alone (15 degrees of freedom), let alone of the
entire hand posture. However, in real hand postures finger joints are
highly correlated so that the available degrees of freedom are by
far not fully used. For instance, the angles of the last two joints
in each finger fulfill for most hand postures approximately the
relation $\theta_3 = (2/3)\theta_2$
since they are driven by the same tendon.
Another simplification is to equate the joint angles $\theta_2$ and
$\theta_1$ that determine the flexion of the fingers.
While this constraint does not exactly correspond to the situation
of the human hand it is a good approximation.
With these constraints, we can represent each hand posture with
10 parameters, which is the same number of parameters as available
with the five 2D coordinate pairs and which, therefore, can be
identified from the 2D data alone.

While the required mapping from finger joints to image locations
is rather straightforward and can be analytically computed, our
system requires the inverse transformation, which is much
harder and cannot be given in closed form. Therefore, we again use
a learning approach at this processing stage.~For each finger,
we train a {\em Parameterized Self-organizing Map} (PSOM), using data 
from the analytically computable forward transform. The PSOM is a
generalization of the well-known Self-organizing map (SOM). It replaces
the discrete lattice of the SOM with a continuous manifold, combined
with the very useful property that it makes available with each
learned mapping automatically also the associated inverse (for details,
cf. \cite{WalRit96}). Fig. \ref{Fig:gridview3col} illustrates the PSOM-generated 
mapping from the 2D-Fingertip positions to the joint-angle space.

\begin{figure}[t!]\centering
\includegraphics[width=0.8\textwidth]{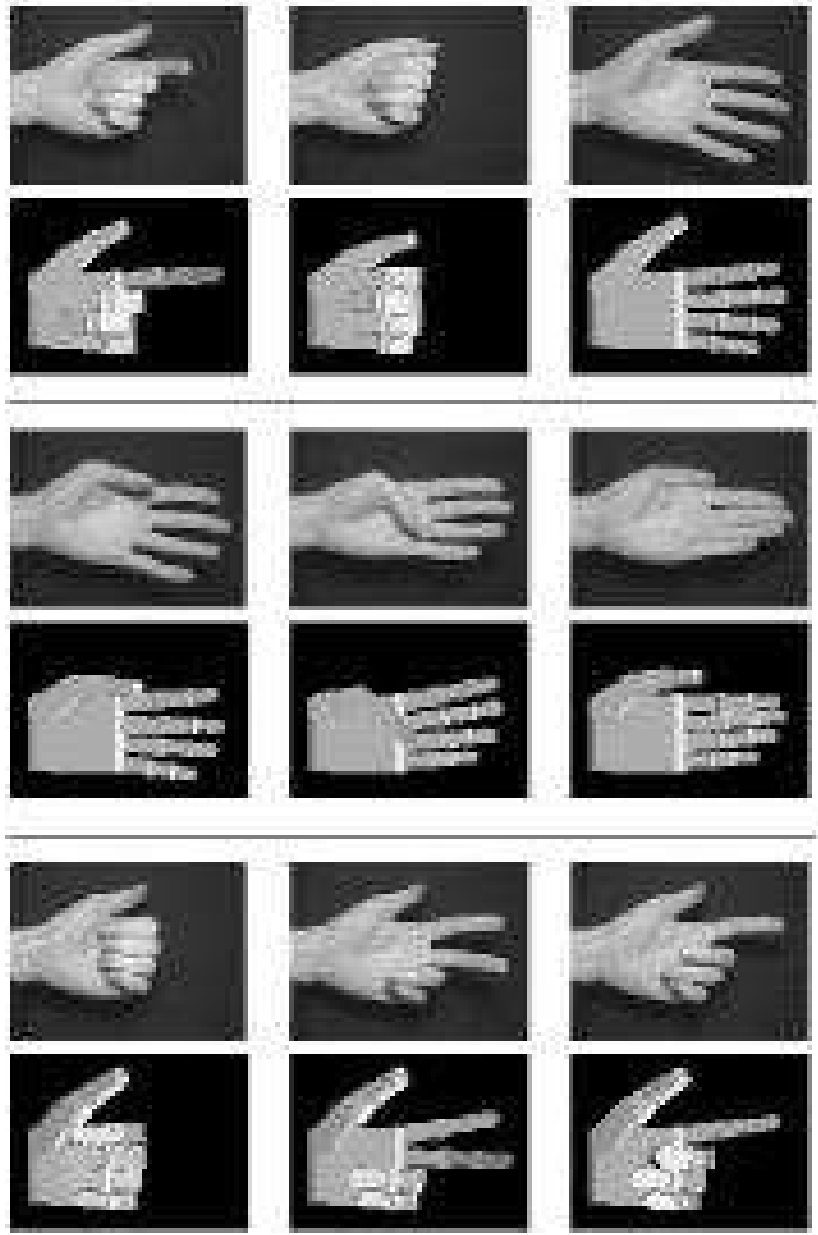}
\caption{Reconstruction results for hand postures with the described  
system.}\label{Fig:kino}
\end{figure}

The development of this system provided us with many insights into
the usefulness of biologically motivated processing principles at
the architectural level of a (still small) vision system.
Fig. \ref{Fig:kino} gives an impression of the system's
accuracy. In contrast
to most other systems with a similar objective, the present
approach can work without special hand markers and makes 
extensive use of learning at several processing levels.

\section{Tactile sensing for robot hand control\label{Sec:haptics}}

Touch is another very important sensory cue that informs us what our
hands do. Although the initial processing steps in the somatosensory cortex
seem to resemble cortical processing in the visual areas (extraction
of local features, such as edges and their motion)\cite{JHT95}, 
one major difference is that the
acquisition of a ``tactile perception'' of an object is a much more interactive 
process than its visual counterpart. In the case of our hands, it is 
intimately connected with a sophisticated control of the mechanical
interaction between their shape-variable sensory surface and the grasped
object. This makes the establishment of computational models an even
bigger challenge than in vision. At the same time, the gulf between the
tactile sensing of our hands and current technical solutions is much
larger than in the case of vision, where rather good camera sensors
are available.

\begin{figure}[h!]\centering
\includegraphics[height=5cm]{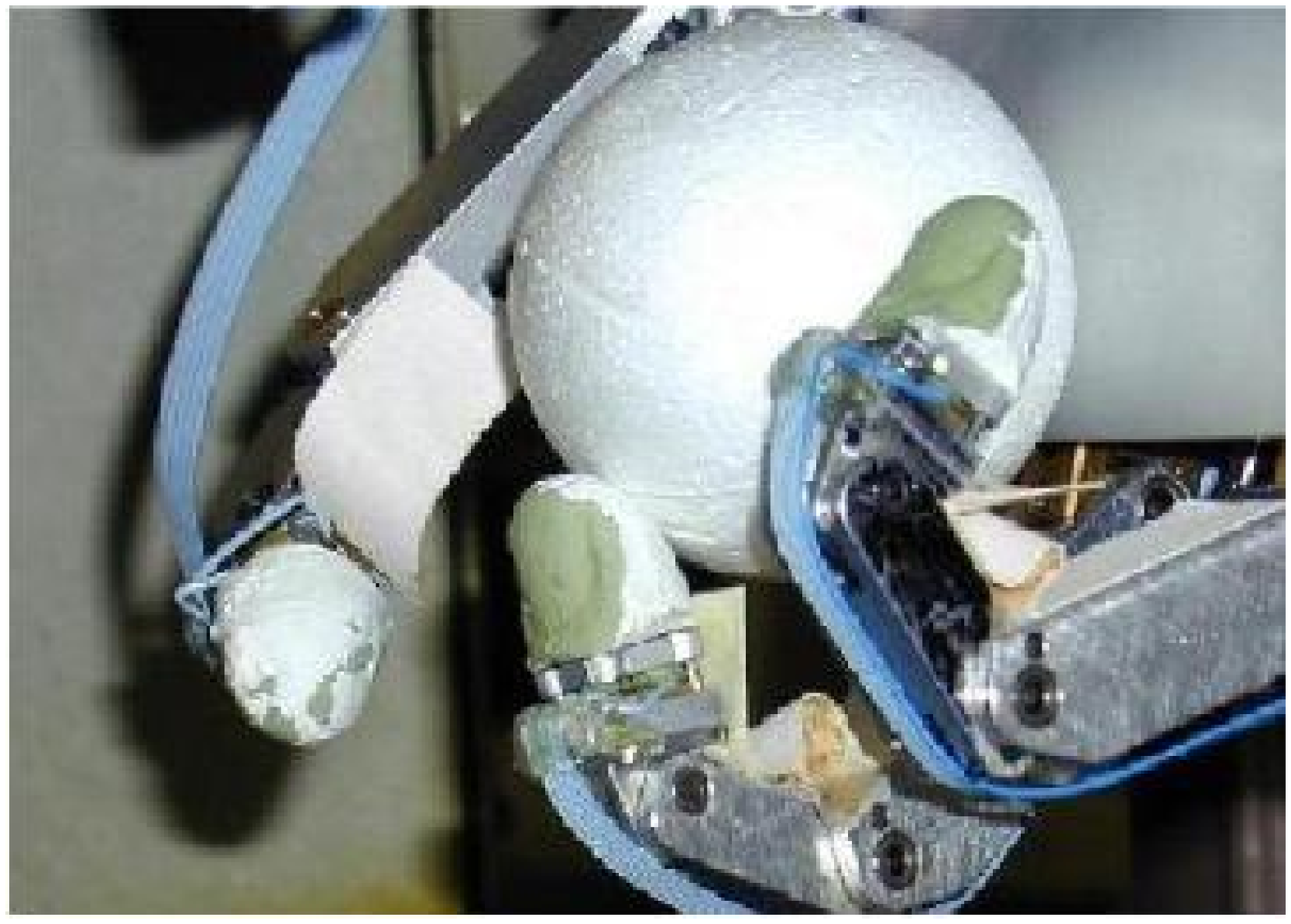}
\includegraphics[height=5cm ]{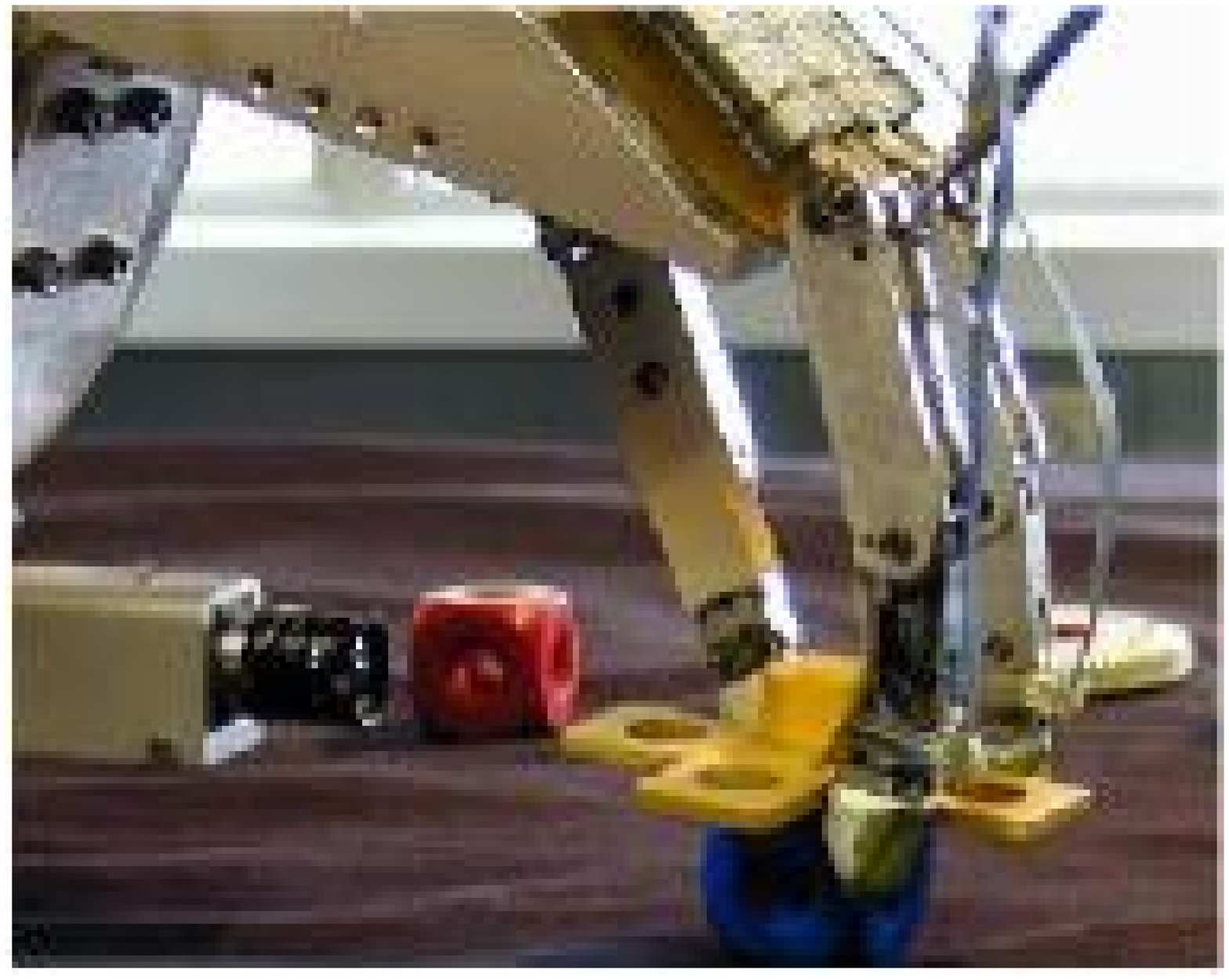}
\caption{Three-fingered hydraulically operated hand for dextrous
grasping.}\label{Fig:TUM-Hand}
\end{figure}

However, its active nature makes the understanding of haptic perception 
a key issue for insights into how the brain controls action. One way to
approach this issue with technical robot systems is to study the control
of multifingered robot hands for haptic discrimination and manipulation
of objects. Fig. \ref{Fig:TUM-Hand} shows a three-fingered robot hand developed at the
Technical University of Munich \cite{TUM} that we have used to explore some of the
issues involved. Each finger is approximately of the size of a human
finger and can be moved about four joints with a total of three 
degrees of freedom (two distal bending joints are coupled, as described
in Sec. \ref{Sec:grefit} for the virtual hand model employed there).
The joints are actuated hydraulically, which allows a fast and forceful
movement control but at the price of hysteresis effects introduced by
friction within the hydraulic cylinders of the actuators. This prevents
the implementation of a reliable joint angle control purely based on
oil pressure data from the actuators and makes additional tactile and
force sensing an even more important necessity.

\noindent\begin{figure}[h!]\centering
\includegraphics[height=4.5truecm]{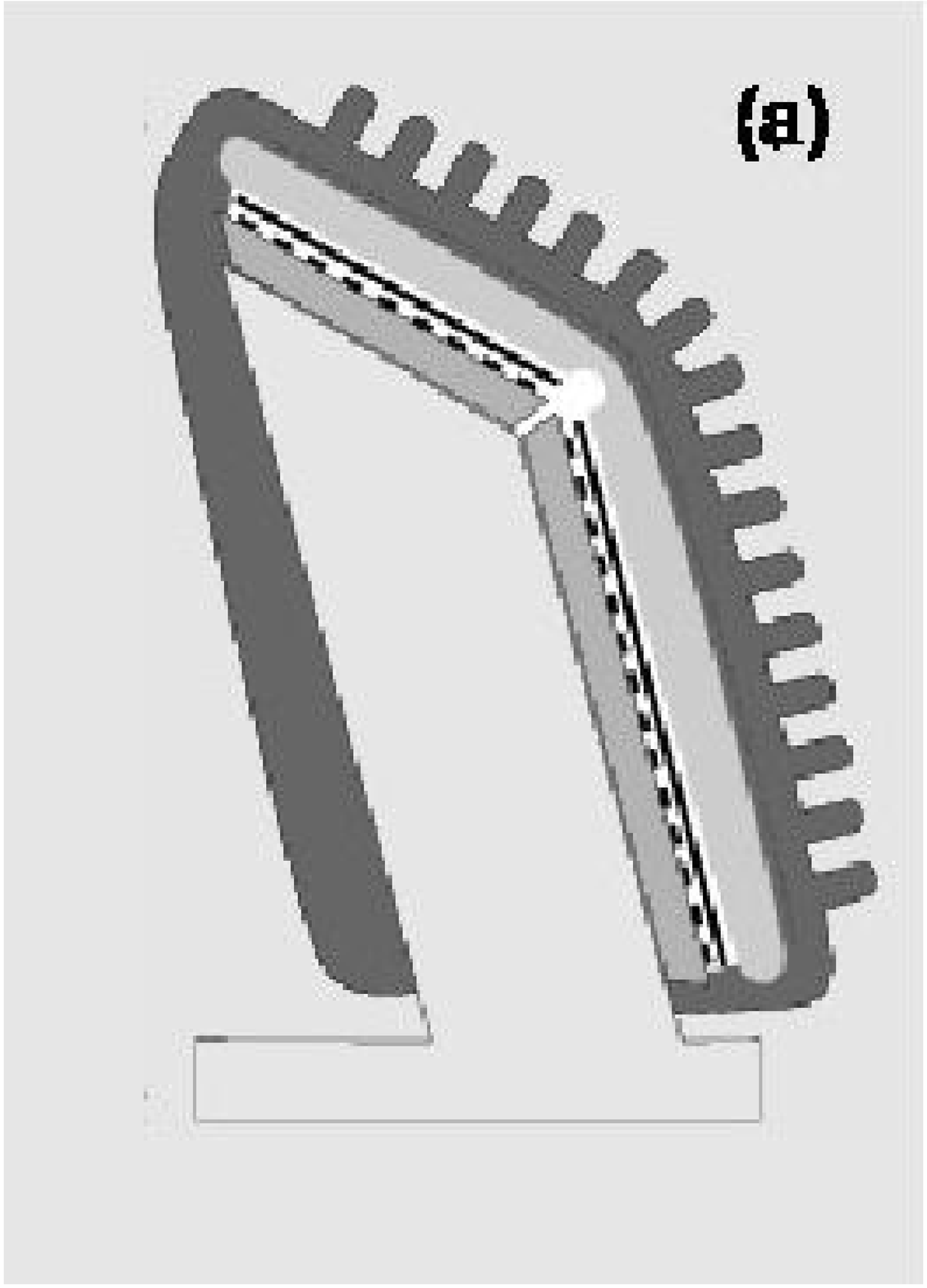}
\hspace{0cm}
\includegraphics[height=4.5truecm]{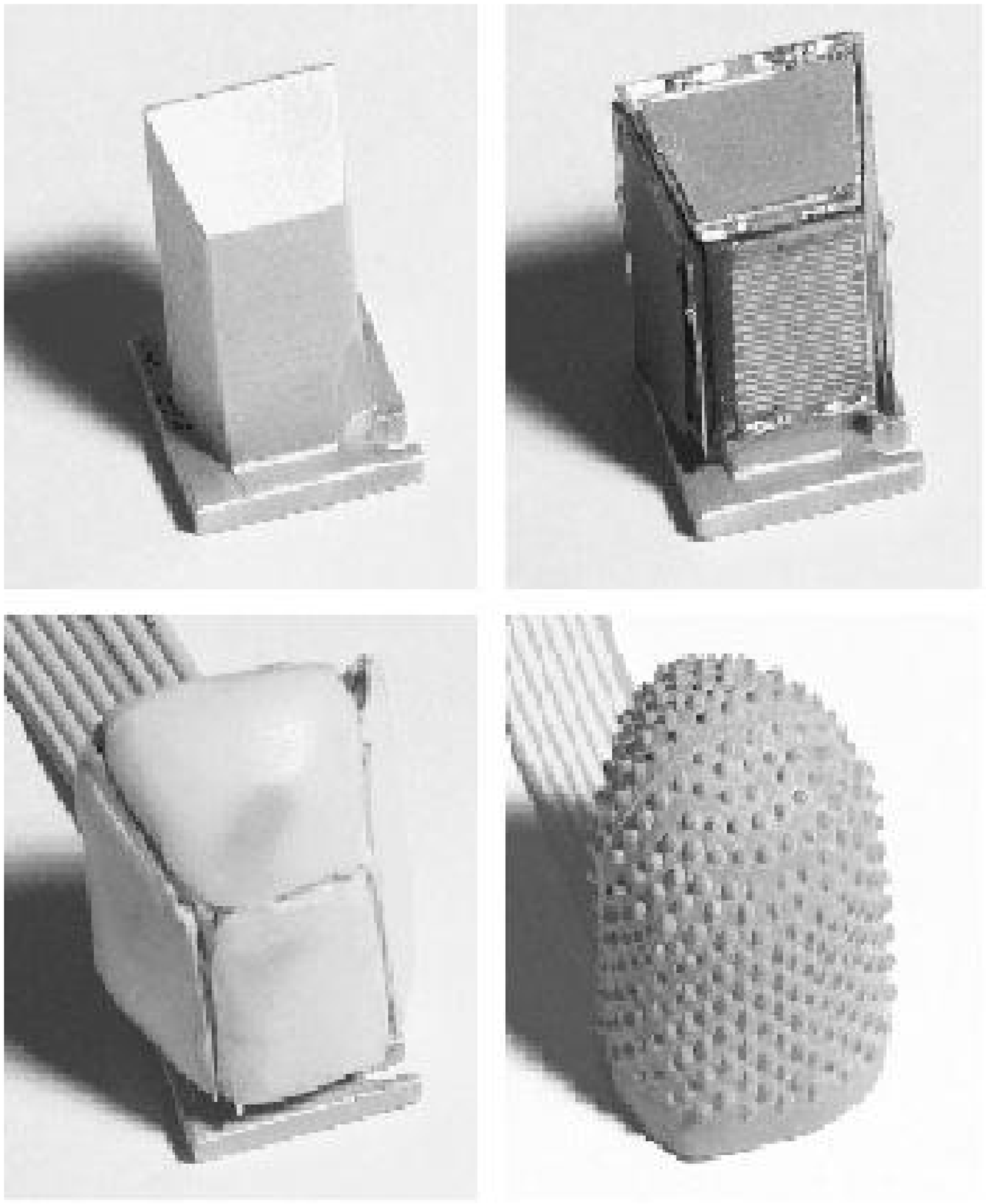}
\hspace{0cm}
\includegraphics[height=4.5truecm]{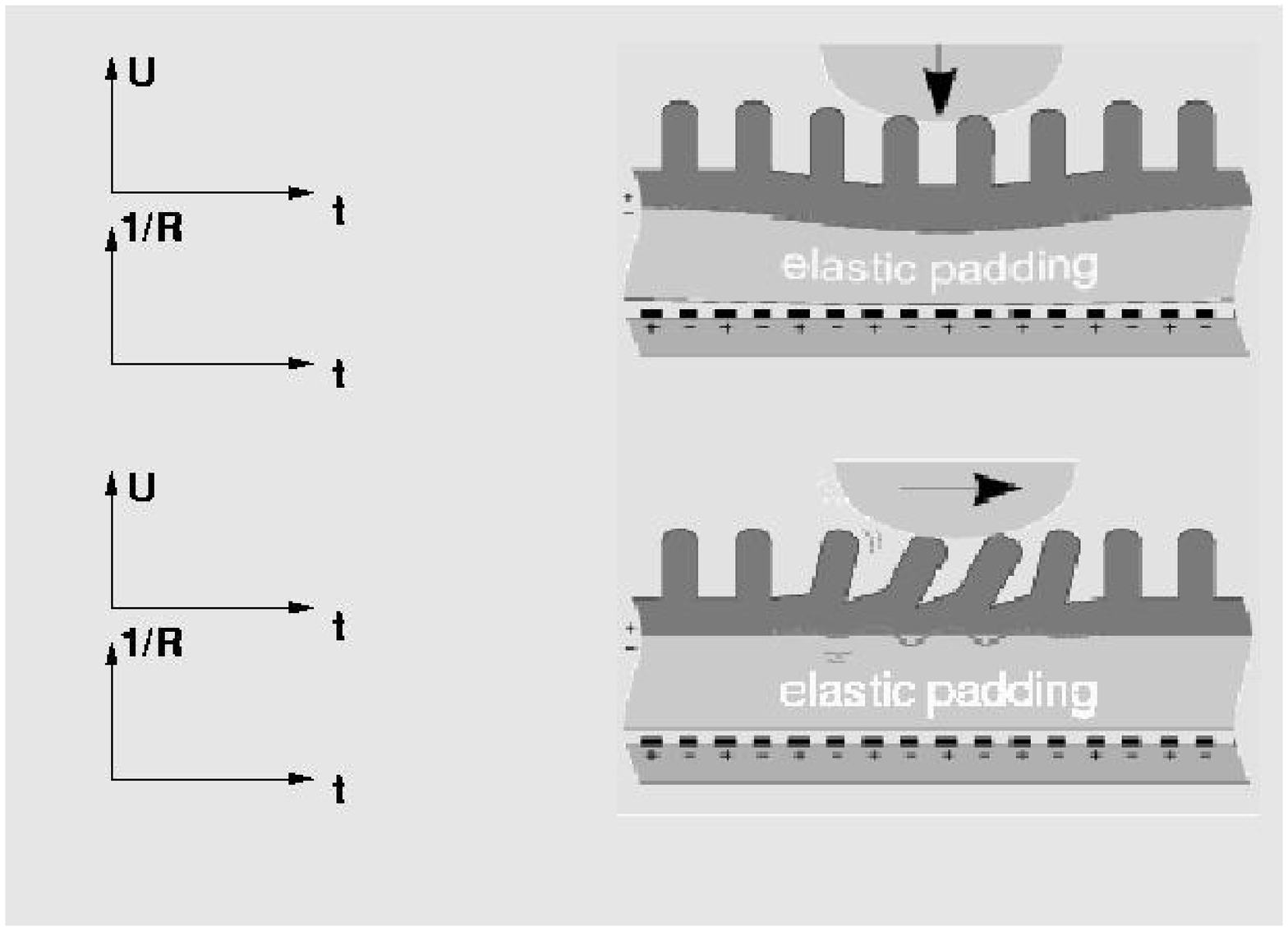}
\caption{Fingertip sensor for detection of pressure and slipping
(from \cite{JockuschWalterRitter1997-ATS})\label{Fig:FingerSensor}}
\end{figure}
To equip this hand with some basic tactile sensing, we
have developed miniaturized finger tip sensors that allow to measure
for each finger tip the force vector caused by an object contact.
An early prototype of the sensors \cite{JockuschWalterRitter1997-ATS} 
used two types of pressure sensitive
foils to mimic the tonic and phasic responses of the skin that are
used to control a static force and to react to rapid force changes
indicating, e.g., object slipping (Fig.\ref{Fig:FingerSensor}). 
Based on experiences with this 
prototype we succeeded in a subsequent, simplified sensor design
(visible in Fig.\ref{Fig:TUM-Hand})
that was much easier to fabricate and to maintain and that
allowed to obtain both tonic and phasic
signals with the use of a single sensor material
only \cite{Jockusch2000-ITM}. 

Obviously, the information gained from these sensors is much more limited
than that provided by a tactile skin covering the entire hand. However, we have seen
in the case of the artificial vision system that the information of finger
tip locations in the visual image carries already a surprisingly large part of the 
information about the 3D hand posture. Similarly, we may hope that the knowledge 
of the finger tip forces can provide us with an important part of the information
necessary to control grasping actions.

\begin{figure}[h!]\centering
\includegraphics[width=0.8\textwidth]{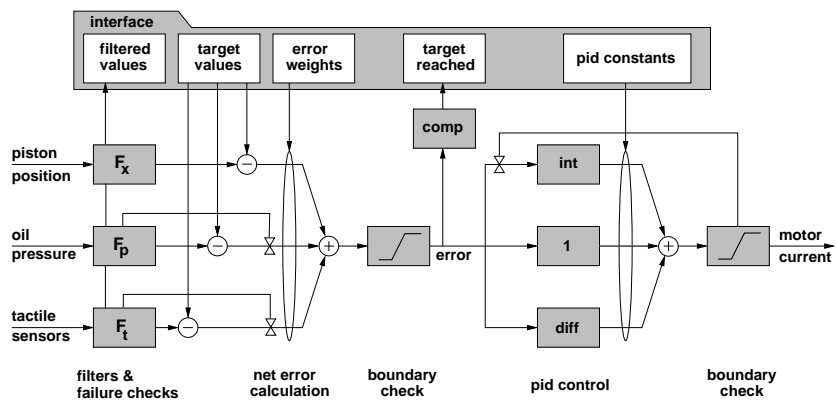}
\hfill
\caption{Control architecture for finger control of the TUM Hand.}
\label{Fig:finger_control}
\end{figure}

Following this assumption we have implemented a control architecture
for basic finger synergies that are involved in dextrous grasping 
(Fig. \ref{Fig:finger_control}). 
The control is based on the minimization of an error function that is
formed from a weighted sum of error signals in different sensory modalities:
a first set of inputs is provided by the force measurements at the finger
tips. A second set of inputs is provided from the oil pressures in the
hydraulic actuators. Finally, a third set of inputs encodes the oil
piston positions in the actuator unit.
By specifying suitable sets of weighting coefficients
for the error contributions we can endow the fingers with different types
of reactive behavior. Switching between such behaviors is achieved with
a finite state automaton whose state transitions become triggered either
by sensory events or by top-down signals and whose states activate
the weighting coefficients for a particular finger behavior.

With this scheme we have implemented several simple grasp primitives,
such as preshaping the hand for grasping an object, closing the fingers
until object contact, maintaining a target force level when holding an
object and reacting with a preset compliance when a human tries to
take the object away.

These primitives permit the robust grasping of ``roundish'' objects; however,
the tactile discrimination of the finger tip sensors is insufficient to 
permit automatic grasps of more complex objects. As a remedy, we have extended
the sensing system in a ``non-biological'' way by adding a hand camera that
evaluates a close view of the object in order to orient the fingers appropriately
for a successful grasp. During grasping itself, this setup permits the combined
use of tactile information from the finger tip sensors with visual information
from the camera. For initial results on the integration of these two
sensory channels to judge grasp reliability, 
see, e.g., \cite{HeidemannRitter2001-VCO}.

\section{Issues of "Real World Learning"}

Although particularly the described hand posture recognition 
system relies heavily on learning algorithms to
attain its capabilities, we think that learning in this as well as in many
similar artificial systems still differs very much from what biological
nervous systems do. We do not suspect that the main reason for this is that
the ``microscopic'' learning rules operating at the level of the formal 
neurons are too far off the mark; they are grounded in well-established
principles such as Hebbian learning and error correction and we would
expect more sophisticated learning rules of the future to differ in detail
but not so much in principle. However, we do suspect that the main difference
lies at the level of the architecture that organizes how and what the
different modules learn and what information they have available.

A first and major difference with biological systems is that learning
is not really integrated with the operation of the system. Instead,
the training phase is separate and training occurs at the level of the
system modules, not at the level of the system as a whole. Both features
are rather typical for most current approaches to exploit learning
for systems that are comprised of several adaptive modules. 

A second difference is the need for carefully prepared
training data sets in which input and target values typically must
be provided in well-defined positions of a data vector (e.g., an array
of 2D finger tip positions for the first and a vector of joint angles
for the second stage of the present system).

These differences make learning still rather artificial and limited.
Real world situations offer no nice separation of operation and
training phase; they also offer no nicely labeled training data vectors
and they give even less opportunity for ``system surgery'' to permit
training of individual modules.

We think that the sketched problems reflect the lack of a good architecture
that specifically supports learning. While the issue of training ``inner''
(or ``hidden'') modules has been addressed quite forcefully in the
context of the multilayer perceptron type approaches, this approach
has faced difficulties with the scaling issues discussed earlier; it is
also well known that multilayer perceptrons are vulnerable to 
``catastrophic interference'' when used for incremental learning, 
a difficulty that is exhibited to a much lesser extent by other,
more ``local'' network models, such as radial basis functions,
local linear maps or other kernel based approaches, such as the 
recently very popular support vector machine \cite{scholkopf98}. 
On the other hand, these models
usually lack the elegant scheme for backpropagating training signals
through entire modules, which made the multilayer perceptron so attractive 
for the construction of modular systems. However, it is well known
that each backpropagation step leads to a significant attenuation of the training
signal, so that this technique again tends to be limited to shallow
architectures that seldom possess more than four hidden layers.
To circumvent these difficulties, researchers have developed an arsenal
of various, often rather heuristic ``tricks'' \cite{Tricks98}; other important
strategies have been a
combination of training and incremental network construction, such
as the cascade correlation architecture \cite{fahl90} that allows
also the use of modules that offer no error
backpropagation mechanism \cite{LittmannRitter1996-LAG}. However,
all these approaches can in our view at best provide a partial 
solution to the above problems.

We suspect that a more severe reason for the limitations of current
learning approaches is an undue emphasis of the view of learning
as a task of identifying an unknown input-output mapping, assuming
input and target values as given. It is this latter assumption that
is almost totally unfulfilled in most real world learning situations.
The paradigm of reinforcement learning \cite{SuBaBook98} tries to address this issue
by weakening the requirements on the available training information
to the extent that the learning system is only informed
(e.g., by a scalar ``reinforcement value'') about relative success
or failure. This leads to conditions that can be met in many real world
situations; at the same time the now greatly impoverished conditions
to acquire useful information make the learning task much harder.
This has prevented reinforcement learning to scale to situations
of realistic complexity, with the exception of a select number of
cases where ingenious encoding of task variables has succeeded 
to focus the search of the reinforcement learning algorithm on
the most promising part of the state space from the outset \cite{mahad92}.

In our view, the more promising way to lift learning to the complexity
of real world situations is not to impoverish the conditions for
information acquisition, as in the paradigm of reinforcement learning,
but instead to put very large efforts into gaining as rich information
from the environment as possible. Such an approach
also seems to be much better in line with what we see in living brains
which obviously devote a large proportion of their processing capacity
to extract useful regularities from a rich spectrum of sensory inputs 
and to actively coordinate the available sensors and processing resources
in order to optimize this process in various ways. Such optimization might 
also require a combined use of supervised and unsupervised learning 
strategies, and it has recently been suggested \cite{Doya99} that
the subdivision of the brain into the neocortex, the cerebellum and
the basal ganglia might reflect an architecture, in which these
three structures provide the substrate for 
unsupervised, supervised and reinforcement learning, respectively.

In the following section
we argue that the main task of such a system, the
extraction of useful regularities from the
environment, shares many goals and computational issues
with the field of {\em datamining} and suggest to consider the
realization of more powerful learning architectures from a
datamining perspective.

\section{A datamining perspective of learning\label{Sec:datamining}}

Datamining is a field that has emerged in the last decade in response
to the needs created by the explosive growth of data acquired in many
fields of science, but also in finance, business enterprises,
communication networks and other areas of daily life \cite{fayyad96,adrians96}. 
Automated data analysis in such situations has become one of the main 
challenges for the future since the explosive growth of our data acquisition 
abilities seems to face no imminent limit.
In particular in the business and finance 
sectors, the early detection of important trends or regularities in
the acquired data can be of vital importance for the survival of a 
company. Yet, the explosive growth of our data acquisition abilities
rules out any solution that relies on human inspection. Therefore,
companies but also scientists find themselves in a position where they
have to develop tools that can autonomously process large collections
of data looking for patterns and regularities that can often only be
vaguely characterized in advance of their detection. 

This situation is
very analogous to that faced by the brains of higher animals: they, too,
are connected to a huge number of sensors that continuously acquire
raw data at a very high rate. Again, the relevant amount of information
in these sensory data constitutes only a tiny fraction of their total
volume and is most often
encoded in subtle patterns that must be detected against a massive
background of noise and irrelevant signal variability. In the case of
living organisms the extremely high survival value of, e.g., recognizing
a predator early even if highly masked in a complex visual or acoustic
background is obvious and we are highly impressed by the superb solutions
crafted by natural evolution in response to these needs. While our own
vision of the world has been created by the same process and imposes on
our imagination a strong bias as to what the world ``is'', we are aware
that our own perception is just a species-specific solution of extracting
from a particular combination of sensory data streams behaviorally relevant
regularities that are mediated to us in categories that are so deeply
engrained into our conscious existence that we have the greatest difficulty of
imagining anything beyond the ``rendering result'' of our own, species specific solution.
Yet we see that brains of other species are connected to sensors that
provide them with access to sensory dimensions totally alien to us,
such as ultrasonic reflections, electric and magnetic field lines,
or polarization of light, to name a few of the cases that have attracted
considerable research.

Therefore, we think that much of the activity of brains is rather aptly
characterized as a sophisticated form of datamining evolved by nature: the rapid 
and highly performant extraction of even subtle regularities from huge amounts 
of raw sensory ``data'' and their representation in stable entities that
are suitable as a basis for rapid decisions about reactions involving
the comparably low-dimensional motor apparatus of an organism.

From this perspective, the task of developing a datamining system 
might be best viewed as the task of building an artificial brain for 
a sensory domain that does not occur in nature but
in the domain of the particular application.
Conversely, one may expect that the search for deeper insights
into processing strategies of real brains might benefit from
experiences and methods developed in the field of datamining.

In fact, there are already many points of contact visible 
between methods employed in datamining and brain modeling \cite{ripley96}:
One major task in both fields is the need for dimension reduction as a
first step to cope with the high bandwidth of incoming data.
This is reflected in the high amount of attention devoted in
both fields to methods such as principal component analysis (PCA),
independent component analysis (ICA), as well as their nonlinear generalizations,
such as self-organizing maps or autoassociator-networks.
Another important issue in both domains is the identification 
of regularities and structures in data of various kinds, putting
cluster algorithms, prototype formation and identification of
manifolds into a shared focus. Further closely related issues
are model extraction, classification and prediction, for which
we have seen the development of neural network based classifiers,
non-linear regression and, more recently, the support vector
machine approach whose neurobiological significance still
awaits clarification. 

It is hard to believe that the prominent presence of these methods
in the two fields is largely coincidental. Instead, we think that
this situation reflects a rather tight relationship between tasks
in both fields that should be exploited in future research. 
In particular, experience from our current work shows us that
the realization of more powerful learning architectures
can benefit from such a program in at least two ways:

\medskip
1. Datamining can contribute many techniques that are well-suited
to create a first layer of representations that reflect important
regularities and invariances of the environment and that offer
more stable features than the sensory data themselves.
In the described hand posture recognition system we have created 
this layer still by design. Its automatic and data-driven construction 
is an essential step to avoid the limiting need for manually prepared 
training data. We were already able to carry out this program to a considerable
extent in a somewhat different domain of discrete object classification
\cite{HeidemannRitter1999-CMN,HeidemannLueckeRitter2000-ASF}, 
using a combination of PCA and clustering, and are currently 
working towards extending these techniques to include cases involving
continuous manifolds.

\medskip
2. Methods for data visualization and -- more recently -- 
data sonification \cite{HermannHansenRitter2001-SOM,HermannRitter1999-LTY}
can help to summarize the inner state of a complex learning system
in various compact ways to facilitate monitoring of the learning
process for a human observer. While this does not directly aid the
learning process itself, it provides the researcher with additional
``windows'' into the learning dynamics that can be very helpful to
gain insights into the interactions that may occur among learning
modules. So far, we have explored this approach for still rather
simple situations \cite{HermannRitter2002-Cryst}; our next target is to apply
this method to the gesture instructed robot system described in the
following sections.
\medskip

However, to view learning from a datamining perspective alone cannot
make disappear the ``curse of dimensionality'' that plagues
any approach that attempts to learn everything from autonomous
and unsupervised exploration. To overcome this problem requires to
offer the learner more useful information than reinforcement
alone can provide \cite{Schaal97}. In the next section we will 
argue that the paradigm
of imitiation learning appears very promising to fill this need. 

\section{Learning as imitation of actions\label{Sec:imitation}}

Learning from a datamining perspective alone still would
confront the learner with the formidable task of having 
to make a large number of difficult
``discoveries'' in a huge search space. 
Everyday experience suggests
that at least human learning is rather different: while we
occasionally learn new things by pure exploration and discovery (which then usually
takes rather long), we acquire a much larger number of skills by
imitation of observed, successful actions of others. While such
``imitation learning'' still is highly non-trivial for reasons more
fully pointed out below, it provides the learner from the outset with
much richer information that is suitable to narrow down search spaces
considerably so that the more general, but weaker learning strategies
of unsupervised or reinforcement learning can provide the now much
smaller, missing information.

The term ``imitation learning'' might suggest that learning that follows
this paradigm now becomes extremely easy. Unfortunately, this is by no means true.
The reasons are at least threefold (\cite{Breazeal00,Mataric00}): $(i)$, usually, the observed
action cannot just be copied ``verbatim'' but must instead be transformed
and adapted for the learner's situation that usually differs in
several respects from that of the ``model''. Related with this
is the problem that $(ii)$ the available input often provides only very partial 
information, e.g., in the case of a hand action the learner may be able
to visually follow the spatio-temporal geometry of the hand movements, 
but will have no sensory access to the tactile sensations and the forces
that accompany (and may be essential for) successful carrying out of the
action. Finally, $(iii)$, the observed action is only available in the
form of ``low level'' sensory inputs embedded in a usually 
complex background of sensory events that are not directly relevant 
to the action of interest. The learner has to do extensive preprocessing
to extract from such input a more concise representation of the observed
action that then can serve as a starting point for solving the remaining
issues $(i)$ and $(ii)$.

\begin{figure}[ht]\centering
\includegraphics[width=0.8\textwidth]{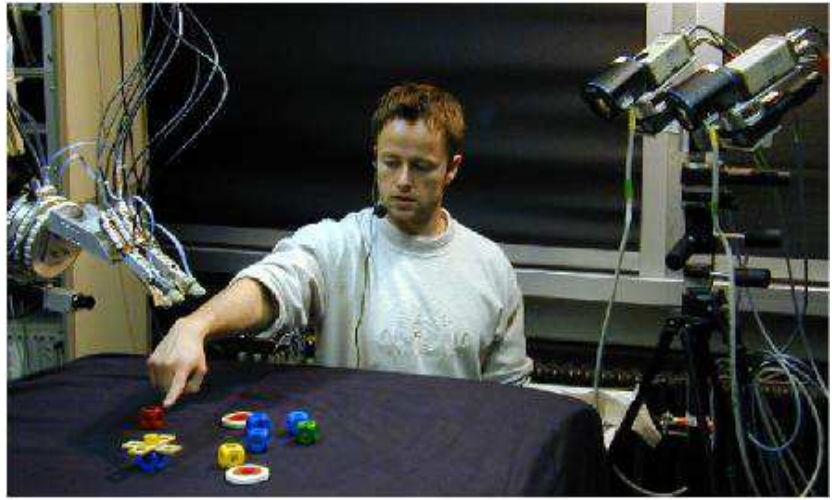}
\caption{Gestural control of robot actions in the GRAVIS system
as a basis scenario towards imitation learning}\label{Fig:patrick}
\end{figure}

To study these issues more concretely, we have chosen the scenario of 
gesture-based control of a multifingered robot hand to carry out 
pick-and-place operations on objects laying on a table and indicated
by manual pointing gestures of a human instructor (Fig \ref{Fig:patrick}).
While this does not yet follow exactly the paradigm of imitation learning
(the robot does not imitate the gestures of the human instructor; instead
it visually observes the gestures and then carries out actions {\em indicated}
by them) this scenario shares many issues that are central
to imitiation learning and the developed solutions will then serve as a 
sound basis that makes proper imitation learning feasable in a next step.

For the realization of GRAVIS (\textbf{G}estural
\textbf{R}ecognition \textbf{A}ctive \textbf{VI}sion 
\textbf{S}ystem robot\cite{SteilHeidemannJockuschRaeJungclausRitter2001-GAF}) 
we had to complement the so far described
functional modules for hand posture recognition (of which the system 
only contains a strongly simplified version) and finger control by
a significant number of further modules for subtasks such as performing
various coordinate transformations (to address $(i)$), robot arm 
and manipulator control (augmenting the approaches sketched in
Sec. \ref{Sec:haptics} to address $(ii)$) and object recognition
(contributing to issues $(iii)$). In the following section, we will 
describe the GRAVIS system more fully. However, for lack of space we will not go
into any details about the involved modules; instead we will concentrate the
discussion on the architecture level, in particular, on the attention
mechanism that allows the system to focus its processing resources on
those parts of the input that are likely to contain useful information
for following the instructor's actions.

\section{Gesture based control of robot actions}

GRAVIS central task is to watch a human instructor for commands that it then
will carry out. GRAVIS is the result of an ongoing, larger scale research effort 
(Bielefeld Special Collaborative Research Unit SFB 360 \cite{RickWachs})
aiming towards the systematic investigation of principles needed to build artificial
cognitive systems that can communicate with a human in an intuitive way,
including the acquisition of new skills by learning. 

As a result of our interest in the role of hands
for understanding sensory-motor intelligence and its links with intuitive,
demonstration-based communication, a major part of the development of GRAVIS
is centered around the recognition, learning and carrying out of hand actions
and gestures in the context of communication. Currently, GRAVIS is able to recognize  
three-dimensional pointing gestures of its instructor and to interprete them
as commands to pick up objects laying around on a table and to deploy
them at positions that again can be designated by pointing gestures.


Obviously, an important element for such a mode of communication is 
the establishment and continuous maintenance of a shared {\em focus of attention} between
robot and human user \cite{scassellati96mechanisms}. 
Since the majority of robot work tasks are 
 related to vision and geometry and in view of the
important role of deictic and spatial gestures in this process,
our robot is equipped with an active stereo camera system to
enable the attention mechanism to shift its attention across a large portion
of the visual scene. 

\begin{figure}[h!]
\centering
\includegraphics[width=0.8\linewidth]{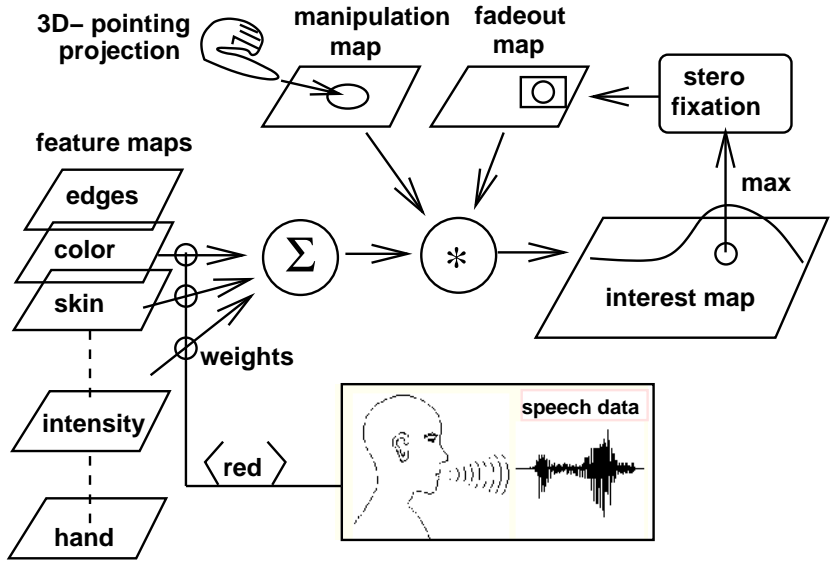}
\caption{The saccade subsystem integrates  spatial aspects of the task by
  summation and multiplication of  spatial   feature maps. }\label{Fig:Attention}
\end{figure}

For the implementation of the attention mechanism we 
use a layered system of topographically organized ``neural'' feature
maps whose task is the integration of different low-level cues into
a continually updated focus of attention (Fig. \ref{Fig:Attention}).
This structure is loosely motivated by  
the current neurobiological picture about sensory integration
mechanisms in the superior colliculus, which is responsible
for targeting visual saccades.  One attractive feature of this
approach is its simple extensibility by additional
layers, allowing a flexible integration of new feature
types into the attention control mechanism. 
Similar approaches, however with 
fewer or less complex maps have been investigated also by \cite{Ferrell96}
in the context of the COG project \cite{BrooksXX} or by \cite{Driscoll98,BreazealScassellati1999,Sethu2001}. 

In more detail,  the vision system computes from the stereo images 
 a  number of feature maps indicating the presence of 
oriented edges, HSI-color saturation and intensity, motion (difference map), 
and skin color.  As one of the main goals of the system is to recognize
and track human hands, we multiply the  difference map (indicating movement) by  the
skin segmentation map (indicating a hand).   The result is a ``moving skin''
map, which is then treated as a feature map in its own right.   
A weighted sum of these feature maps is multiplied  coordinate-wise with a further
manipulation map and a fadeout-map to form the final attention map.

After a Gaussian smoothing to favor small saccades, the location of maximal activity in the
attention map is used to define the next fixation point.   The multiplicative
nature of the  manipulation map allows one to direct (or suppress) fixations 
into (or within) whole regions according to the availability of top-down 
information (cf. below).
Finally, the fade-out map has the task to suppress activity in recently visited 
regions and also in areas that would command fixations that are incompatible
with the joint limits of the camera head.  However, 
in order to maintain interest in pointing gestures, the ``moved skin'' 
 map is always additively superimposed to the fade-out map. 
As a last step, the location of maximal activity in the attention map
is centered on the nearest object in its vicinity, and a stereo matching
algorithm is used to estimate depth and to obtain a stereo correction 
that centers both cameras on the new fixation point \cite{JungclausRaeRitter1999-AIS}.

\begin{figure}[h]
\centering
\includegraphics[width=0.8\linewidth]{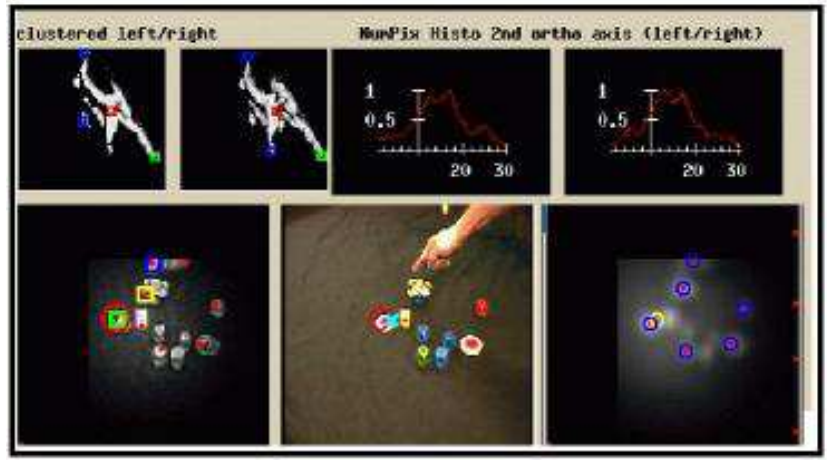}
\caption{A pointing gesture is evaluated in several steps, using skin segmentation
  (upper left)  and mult-layer perceptrons to classify pointing (upper right).
 A  corresponding restriction of attention to a region of interest is generated and
  projected on the table, here indicated by the square in the lower middle
 display. Other points of interest in the raw image (left) and after stereo
 matching (right) are further candidates for fixations in case any pointing
 information is available.}
\label{Fig:Point}
\end{figure}
The weighted summation of the topographic maps provides also a
very convenient way for  top-down propagation of information from higher
processing levels to the perceptual level. For instance, 
interaction with the human user can modify the attention map  by two 
different mechanisms. If the  hand and gesture recognition system
detects a pointing gesture in the image,  the 3D-direction of the pointing 
finger is  computed and its 2D projection into the viewing
plane is used to define a sector-shaped ``pointing cone'' emanating
from the pointing finger into the scene. The pointing cone is represented as
correspondingly localized activity in the ``manipulation map'' layer that
then is multiplied point-wise with the attention map to restrict the 
explorative attention of the next step to the area of the pointing cone
(Fig.~\ref{Fig:Attention}). Additionally,  
if a spoken instruction references a colored object (``
... the red cube ...'') the corresponding weight of the corresponding 
color map is increased to bias the
attention system towards red  spots in the image.
This  increases the probability for fixations on red blobs, but after some 
time a  decay mechanism drives the weighting back to a default
level. 

\begin{figure}[t!]
\centering
\includegraphics[width=0.8\linewidth,height=0.6\textheight]{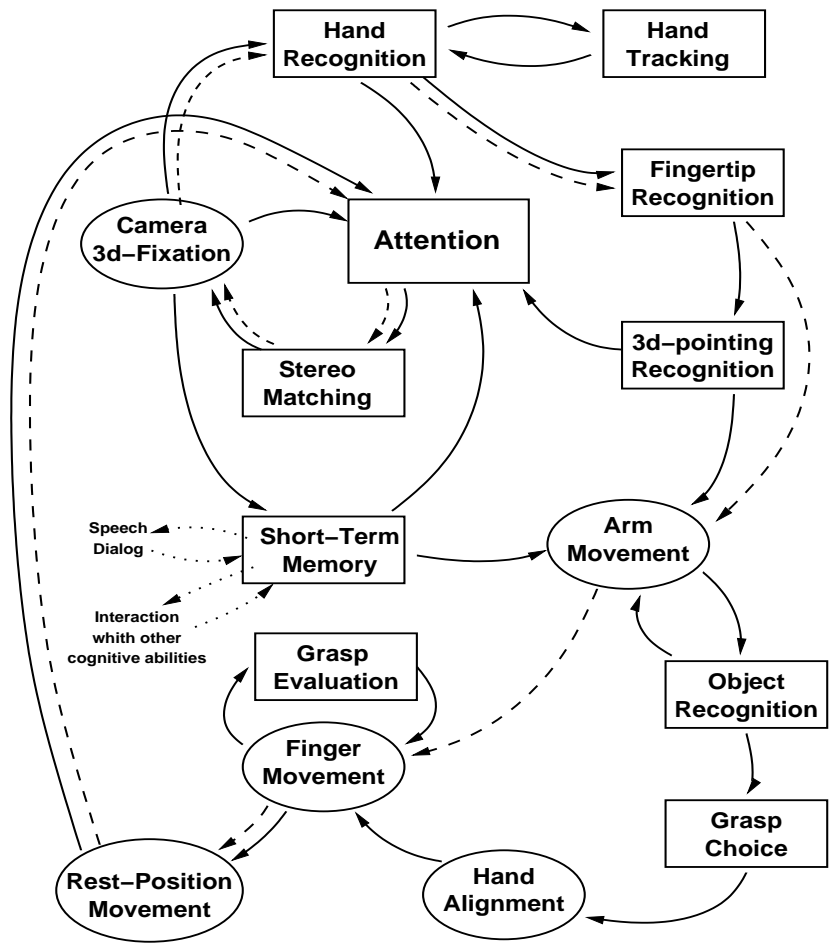}
\caption{The functional modules of the GRAVIS architecture, blocks depict
  software modules, circles hardware control, and arrows the (simplified) 
control flow.  The dashed line shows the behavior sequence for  object
 deployment starting from the rest-position in the lower left corner.}
\label{Fig:Modules}
\end{figure}

While the attention system is a key component to coordinate the activity
of the primarily visuo-spatially directed modules of the GRAVIS system,
the entire task of gesture-controlled pick-and-place operations requires 
participation of a considerable number of further perceptual and motor
basis skills. Fig.~\ref{Fig:Modules} provides a simplified overview of these and
of their mutual interactions. Besides the already mentioned visuo-spatial
modules for hand recognition, hand tracking, fingertip recognition,
3D pointing recognition and stereo matching, additional modules have to
deal with the motor control of 3D-fixations of the stereo camera system,
recognition of the target object and its orientation, a corresponding 
grasp choice and pre-shaping of the hand, the control of robot arm and 
finger movements, and, finally, the evaluation of the attained grasp 
based on feedback from the finger sensors.

To organize the 
control flow between the modules implementing these
skills, which represent  primarily motor-directed aspects of the robot's 
behavior, we use finite-state automata  that can activate or de-activate 
 mutually exclusive behavior modules. Several of these modules
are again implemented by using artificial  neural networks, can
 adapt to changing characteristics of
the task,  and are realized in a hierarchical fashion
similar to the hand recognition system described above.

At the software level, the entire system is implemented as a larger number of separate
processes running in parallel on several workstations and
communicating with a  distributed message passing  communication system (DACS
\cite{Fink95:DACS}) developed earlier for the purpose of this project. 

At the hardware level, the current system comprises
the already mentioned binocular camera head equipped with two  color-cameras
and motorized lenses \cite{AUC} with a total of 10 DOF's and the three-fingered
robot hand described in Sec. \ref{Sec:haptics}.
The hand is mounted on a  PUMA 560 manipulator 
with six DOFs operated in real time and is additionally equipped with 
a wrist camera to  view  the end-effector region.

\section{Discussion} 

Usually, systems of the complexity of the described GRAVIS system
or beyond are research prototypes that are rather far from a
portable product that can be easily set up in the same manner
in a different laboratory. At the same time, these systems are still
extremely simple from a biological point of view.

So an important issue to address is: what
insights -- beyond experience -- can be gained from the construction 
of such prototypes?

A first answer to that question is that systems of the kind of GRAVIS
allow us to experimentally explore the interactions of
a collection of sensory, memory and motor skills in the context
of real-world tasks instead of highly simplified and artificial
toy problems that are not characteristic of the tasks brains
were evolved for. Such experiments may give clues about processing
strategies that might otherwise be very difficult to obtain.

For instance, the exploration behavior generated by the attention module 
tends to fixate repetitively upon the most
interesting  points, which are in most cases objects. Here we find a typical 
``emerging regularity'' grounded in a perception-action loop which 
 can for instance be  used  to establish a visual memory by temporal integration
which  stabilizes only the most salient points. Though in this case the regularity 
is rather obvious and detectable in the chain of 3D-fixations alone, 
the availability of a technical system that is open to any kind of
``surgery'' in conjunction with the datamining techniques discussed
in Sec. \ref{Sec:datamining} offers the chance to detect also more subtle 
patterns, implicit dependencies and mutual relationships between 
sensory signals, their intermediate representations, and behavioral states
or actions that can point to less obvious ``emergent'' processing
strategies. If we consider a  more complex chain of events in
 Fig. \ref{Fig:Modules}, for instance a repetitive pointing to objects,
the regularities   will typically not be manifest and easily detectable on the
 hardware level of sensory signals or the stage of a single module alone. Then 
to endow the system with  an efficient learning architecture requires to 
collect information from different hierarchical levels and many  modules, to 
generate useful combined representations and to define appropiate learning on 
several stages from calibration of hardware-near feedback loops up to high-level
organisation of the interactions between  functional  modules. Though there  remain
many questions in integrating these levels, we find that the attentional mechanism
discussed above provides one very flexible  approach to generate
as well regularities to be exploited by higher level modules as  a way to
propagate  top-down higher level knowledge to modulate the lower level
processing. As the human brain compared to our robotic system is much more  complex,
 we  would expect that its cognitive abilities as well rely on a multi-scale
 hierarchy of processing and thus their understanding will require explanations
reaching from  the  level of neural connectivity  up to  the level of interrelations of
functional modules and, as a key element, the organisation of their mutual
relationships. 

%
A second answer is connected with an important issue that
is totally absent in small systems: the issue of {\em scaling}. 
While simple tasks invite many
workable approaches, their number gets drastically reduced when
task requirements increase. For instance, in earlier work we found
that the 3D postures of computer-rendered hand configurations
could be rather accurately predicted with a single, ``monolithic''
neural net that was trained with a sufficient number of computer
generated pictures as input and the known joint angles as output
values. However, we found that we were unable to generalize this
pure ``black-box'' approach to the identification of hand postures of real human
hands. Initial attempts rapidly revealed that the visual variability
of a real hand was much larger than that of the computer hand model
used in previous work. Contributing factors were not only the higher
shape variability, but also visual effects such as shadows or 
variability in lighting that were not modeled in the artificial
images. Another factor was the more limited amount of training data
that could be produced: while synthetic images can be generated
in almost unlimited amounts, it becomes difficult to prepare more
than, say, a few thousand real hand images for training.
Finally, for a computer rendered hand image, all joint angles
are known by construction while for real images such information
is absent.

All these factors were more or less directly related to issues of scaling: 
the transition from computer rendered to real images involved 
a significant up-scaling of image variability; at the same time, 
size of the available training corpus was scaled downward and
in addition the learning algorithm had to be adapted to be 
workable for much more restricted training information 
(2D finger tip locations instead of 3D joint angles).

The higher shape variability, together with the more limited 
available training information led to the necessity of introducing an
intermediate representation at the 2D level (in the present example 
the finger tip locations) from which then the 3D information is
obtained only by a separate stage. The task of this intermediate
representation was twofold: $(i)$ to represent the  
sensory input with a set of features (in the present example 
the 2D finger tip locations) that could be effectively correlated 
with available training information (marked 2D locations in training 
images) so that one can construct the representation
by data-driven learning. $(ii)$ to reduce the task complexity
for the remaining stages by ``condensing'' the task-relevant
information in a set of features that are stable with respect
to non-relevant changes in the sensory input.

A third answer, closely related with the issue of scaling,
is that only the construction of larger systems provides us with
sufficiently realistic opportunities to explore architectural
principles for the organization of a larger collection of
functional modules and their communication. While the implementation
of GRAVIS required in many places also the use of handcrafted
heuristics to make things work, there were a number of
more general principles that proved valuable in many situations:

$(i)$ the adaptive, data-driven construction of 
feature mappings, using neural networks that combined
elements of both local and global feature encoding.

$(ii)$ the use of feature maps to
encode low-level visuo-spatial information 
in a format that is flexible for shaping interactions and
for later addition of new modules.

$(iii)$ a ``vertical'' organization into
several, hierarchically organized processing stages
in which each stage has the more modest goal of a cautious
narrowing down of the solution space instead of attempting
an early final result (for instance, first
identify finger tip regions, then
in each region location candidates, only then final
location).

$(iv)$ a ``horizontal'' organization into  several processing 
streams that were
directed at the same goal (e.g., identify important image locations)
but that used different computational strategies
(neural network based prediction, boundary detection, 
pattern template) proved efficient for achieving robustness 
in an extensible way.

To what extent can work as reported here be useful in neuroscience?
Certainly, we are fully aware that work such as reported here can make no direct 
contribution to modeling the biological brain structures themselves, 
since systems as the above provide no direct correspondences to
experimentally observable neural structures. We would not even claim
that there is any close correspondence in terms of functional units
or their interaction patterns. However, we think that approaches of
the kind as described here can be useful to get an impression of the
computational challenges of problems solved by the brain, and we
can explore the feasability of general computational strategies that 
underlie current hypotheses about processing in the brain at a system 
level. 

For example, by building systems that can extract the 3D posture 
of hands from images, we can study how different representational schemes
(e.g., holistic versus hierarchical) can cope with the problem of
representing information about a complex articulated shape.
The implemented system then permits to study how strategies,
such as the use of multiple processing streams, can be exploited
to increase robustness. By employing learning algorithms to train 
the artificial system we can get upper bounds on the required number 
of training views in order to achieve a particular recognition accuracy. 
The building of actual systems also confronts us with limitations
of current learning approaches, e.g., their need of labelled training 
examples for efficient learning, often in conjunction with ''system surgery'' 
to train the individual modules, and provides us at the same time with
an experimental platform to explore ways to overcome these limitations
and to develop --- in the long run --- learning architectures that
come in their abilities closer to those of living systems. 

Since the abilities of self-generated actions, as well as the imitation
of observed actions are very likely to be crucial for most forms
of real-world learning, we also need to build interactive robot
systems, such as exemplified by the GRAVIS system reported here.
Here too, we illustrated how such systems permit to study biologically motivated
computational strategies, such as the use of topographic activity
maps for attention control, or behavior-based architectures 
consisting of interconnected networks of basis behaviors.
Again, individual modules in our system (as depicted in Fig. \ref{Fig:Modules})
cannot be matched directly onto putative counterparts in the
brain. Despite the absence of such a correspondence a system
such as GRAVIS permits us to explore the feasability of particular
computational approaches --- in this case, a map-based attention
control within a behavior-based architecture
--- that have been inspired by current ideas about how action chains
might be controlled in a neural system. In this way, robot systems
can provide us with a useful ''scratchpad'' for better assessing the
workability of ideas about how a complex perception-action system
might achieve the observed, highly flexible coordination of its
sensing and acting capabilities. Thus, they can aid us towards a
clearer picture about {\em sufficient conditions} to generate particular
capabilities.

Last, but not least, it may be comforting to find that 
architectural principles that are in line with current views of brain function 
can prove valuable for the solution of demanding
tasks of machine perception. Thus, research prototypes of such systems
can offer additional avenues to explore some of the strengths and
limitations of such architectures at a more abstract level and in
ways that might not be feasible with real brains.

\bibliographystyle{plain}

\section*{Acknowledgement}

Part of this research was funded by the German Science Foundation
(DFG CRC 360).

\bibliography{iros02,network}

\begin{thebibliography}{10}

\bibitem{adrians96}
P.~Adrians and D.~Zantinge.
\newblock {\em Datamining}.
\newblock Addison Wesley, 1996.

\bibitem{Bakker96}
P.~Bakker and Y.~Kuniyoshi.
\newblock Robot see, robot do: An overview of robot imitation.
\newblock In {\em Proc. AISB workshop Learning in Robots and Animals}, pages
  3--11, Brighton, UK, 1996.

\bibitem{BreazealScassellati1999}
C.~Breazeal and B.~Scassellati.
\newblock A context-dependent attention system for a social robot.
\newblock In {\em Proc.IJCAI99}, pages 1146--1151, Stockholm, Sweden, 1999.

\bibitem{Breazeal00}
C.~Breazeal and B.~Scassellati.
\newblock Challenges in building robots that imitate people.
\newblock In K.~Dautenhahn and C.~Nehaniv, editors, {\em Imitation in Animals
  and Artifacts}. MIT Press, in press.

\bibitem{BrooksXX}
R.~A. Brooks, C.~Breazeal, M.~Marjanovic, B.~Scassellati, and M.~M. Williamson.
\newblock The {COG} project: Building a humanoid robot.
\newblock In C.~L. Nehaniv, editor, {\em Computation for Metaphors, Analogy and
  Agents}, LN AI 1562. Springer, 1999.

\bibitem{AUC}
H.~I. Christensen.
\newblock The {AUC} robot camera head.
\newblock In {\em SPIE Applications of Artificial Intelligence X: Machine
  Vision and Robotics}, pages 26--33, 1992.

\bibitem{Dau80}
J.G. Daugman.
\newblock Two-dimensional spectral analysis of cortical receptive field
  profiles.
\newblock {\em Vision Research}, 20:847--756, 1980.

\bibitem{Doya99}
Kenji Doya.
\newblock What are the computations of the cerebellum, the basal ganglia, and
  the cerebral cortex.
\newblock {\em Neural Networks}, 12:961--974, 1999.

\bibitem{Driscoll98}
J.~A. Driscoll, R.~A.~Peters II, and K.~R. Cave.
\newblock A visual attention network for a humanoid robot.
\newblock In {\em Proc. IROS 1998}, Victoria, B.C., October 12-16 1998.

\bibitem{fahl90}
S.~E. Fahlman and C.~Lebiere.
\newblock The cascade-correlation learning architecture.
\newblock In D.~S. Touretzky, editor, {\em Advances in Neural Information
  Processing Systems}, volume~2, pages 524--532, Denver 1989, 1990. Morgan
  Kaufmann, San Mateo.

\bibitem{fayyad96}
U.~Fayyad, G.~Piatetsky-Shapiro, P.~Smyth, and R.~Uthurusamy.
\newblock {\em Advances in Knowledge Discovery and Data Mining}.
\newblock MIT Press, Cambridge, MA, 1996.

\bibitem{Ferrell96}
Cynthia Ferrell.
\newblock Orientation behavior using registered topographic maps.
\newblock In {\em From Animals to Animats: Proceedings of 1996 Society of
  Adaptive Behavior}, pages 94--103, Cape Cod, Massachusetts, 1996.

\bibitem{Fink95:DACS}
G.A. Fink, N.~Jungclaus, H.~Ritter, and G.~Sagerer.
\newblock A communication framework for heterogeneous distributed pattern
  analysis.
\newblock In {\em Int. Conf. on Algorithms and Architectures for Parallel
  Processing}, pages 881--890, Brisbane, 1995.

\bibitem{HeidemannLueckeRitter2000-ASF}
G.~Heidemann, D.~L\"ucke, and H.~Ritter.
\newblock A system for various visual classification tasks based on neural
  networks.
\newblock In {\em 15th International Conference on Pattern Recognition},
  volume~1, pages 9--12, Barcelona, 2000.

\bibitem{HeidemannRitter1999-CMN}
G.~Heidemann and H.~Ritter.
\newblock Combining multiple neural nets for visual feature selection and
  classification.
\newblock In {\em ICANN 99, Ninth International Conference on Artificial Neural
  Networks}, pages 365--370, 1999.

\bibitem{HeidemannRitter2001-VCO}
G.~Heidemann and H.~Ritter.
\newblock Visual checking of grasping positions of a three-fingered robot hand.
\newblock In H.~Bischof G.~Dorffner and K.~Hornik, editors, {\em Proc. ICANN
  2001}, pages 891--898. Springer-Verlag, 2001.

\bibitem{HermannHansenRitter2001-SOM}
T.~Hermann, M.~H. Hansen, and H.~Ritter.
\newblock Sonification of markov chain monte carlo simulations.
\newblock In Zacharov Hiipakka and Takala eds., editors, {\em Proc of 7th
  International Conference of Auditory Display, 2001}, pages 208--216, Espoo,
  Finland, 2001.

\bibitem{HermannRitter1999-LTY}
Thomas Hermann and Helge Ritter.
\newblock Listen to your data: Model-based sonification for data analysis.
\newblock In G.~E. Lasker, editor, {\em Advances in intelligent computing and
  multimedia systems, Baden-Baden, Germany}, pages 189--194. Int. Inst. for
  Advanced Studies in System research and cybernetics, 1999.

\bibitem{HermannRitter2002-Cryst}
Thomas Hermann and Helge Ritter.
\newblock Crystallization sonification of high-dimensional datasets.
\newblock In {\em Proc of the Int. Conf. on Auditory Display}, pages 76--81,
  Kyoto, Japan, 2002. Int. Community for Auditory Display.

\bibitem{Jockusch2000-ITM}
J{á}n Jockusch.
\newblock {\em Exploration based on neural networks with applications in
  manipulator control}.
\newblock PhD thesis, University of Bielefeld, Faculty of Technology, 2000.
\newblock http://archiv.ub.uni-bielefeld.de/disshabi/technik.htm.

\bibitem{JockuschWalterRitter1997-ATS}
Ján Jockusch, Jörg Walter, and Helge Ritter.
\newblock A tactile sensor system for a three-fingered robot manipulator.
\newblock In {\em Proc. Int. Conf. on Robotics and Automation (ICRA-97)}, pages
  3080--3086, 1997.

\bibitem{JHT95}
Kenneth~O. Johnson, Steven~S. Hsiao, and I.~Alexander Twombley.
\newblock Neural mechanisms of tactile form recognition.
\newblock In Michael~S. Gazzaniga, editor, {\em The Cognitive Neurosciences},
  pages 253--267. MIT Press, 1995.

\bibitem{JungclausRaeRitter1999-AIS}
Nils Jungclaus, Robert Rae, and Helge Ritter.
\newblock An integrated system for advanced human-computer interaction.
\newblock In {\em UCSB-Workshop on Signals and Images (SIPL)}, pages 93--97,
  University of California, Santa Barbara, USA, 1998.

\bibitem{kaelbling96reinforcement}
Leslie~Pack Kaelbling, Michael~L. Littman, and Andrew~P. Moore.
\newblock Reinforcement learning: A survey.
\newblock {\em Journal of Artificial Intelligence Research}, 4:237--285, 1996.

\bibitem{Kuniyoshi94}
Y.~Kuniyoshi, M.~Inaba, and H.~Inoue.
\newblock Learning by watching: extracting reusable task knowledge from visual
  observation of human performance.
\newblock {\em IEEE. Trans. Robotics Automation}, 10(6):799--822, 1994.

\bibitem{LittmannRitter1996-LAG}
E.~Littmann and H.~Ritter.
\newblock Learning and generalization in cascade network architectures.
\newblock {\em Neural Computation}, 8(7):1521--1539, 1996.

\bibitem{mahad92}
Sridhar Mahadevan and Jonathan Connell.
\newblock Automatic programming of behavior-based robots using reinforcement
  learning.
\newblock {\em Artificial Intelligence}, (55):311--365, 1992.

\bibitem{Mataric00}
Maja~J. Mataric, Odest~C. Jenkins, Ajo Fod, and Victor Zordan.
\newblock Control and imitation in humanoids.
\newblock In {\em AAAI Fall Symposium on Simulating Human Agents}, North
  Falmouth, MA, Nov. 3-5 2000.

\bibitem{MeinickeRitter2001-RBC}
Peter Meinicke and Helge Ritter.
\newblock Resolution-based complexity control for {Gaussian} mixture models.
\newblock {\em Neural Computation}, 13(2), 2001.

\bibitem{TUM}
R.~Menzel, K.~Woelfl, and F.~Pfeiffer.
\newblock The development of a hydraulic hand.
\newblock In {\em Proc. 2. Conf. Mechatronics and Robotics}, pages 225--238,
  1993.

\bibitem{MeyRit92b}
A.~Meyering and H.~Ritter.
\newblock Learning to recognize 3d-hand postures from perspective pixel images.
\newblock {\em Artificial Neural Networks}, 2:821--824, 1992.
\newblock Elsevier Science Publishers B.V., North Holland.

\bibitem{Young94}
GAPC~Burns MP~Young, JW~Scannell and C~Blakemore.
\newblock Scaling and brain connectivity.
\newblock {\em Nature}, 369:449--450, 1994.

\bibitem{NoeRit02}
Claudia N\"olker and Helge Ritter.
\newblock Visual recognition of continuous hand postures.
\newblock {\em IEEE Transactions on Neural Networks}, May 2002.

\bibitem{Tricks98}
Genevieve~B. Orr and Klaus~Robert M\"uller, editors.
\newblock {\em Neural Networks: Tricks of the Trade}.
\newblock Springer, Berlin, 1998.

\bibitem{Perret89}
D.I. Perret, M.U. Harries, R.~Devan, S.~Thomas, P.J. Benson, A.J. Mistlin, A.J.
  Chitty, J.K. Uietanen, and J.F. Ortega.
\newblock Frameworks of analysis for the neural representation of animate
  objects and actions.
\newblock {\em Journal of Experimental Biology}, 146:87--113, 1989.

\bibitem{Peters00}
Gabriele Peters.
\newblock Theories of three-dimensional object perception --- a survey.
\newblock In {\em Recent Developments in Pattern Recognition (Part I)},
  volume~1, pages 179--197. Transworld Research Network, 2000.

\bibitem{RickWachs}
G.~Rickheit and I.~Wachsmuth.
\newblock Collaborative research centre "situated artificial communicators" at
  the {U}niversity of {B}ielefeld, {G}ermany.
\newblock {\em Artificial Intelligence Review}, 10(3-4):165--170, 1996.

\bibitem{ripley96}
B.~D. Ripley.
\newblock {\em Pattern Recognition and Neural Networks}.
\newblock Cambridge University Press, Cambridge, Great Britain, 1996.

\bibitem{scassellati96mechanisms}
B.~Scassellati.
\newblock Mechanisms of shared attention for a humanoid robot.
\newblock In {\em Embodied Cognition and Action: Papers from the 1996 AAAI Fall
  Symposium}, 1996.

\bibitem{Schaal97}
Stefan Schaal.
\newblock Learning from demonstration.
\newblock In M.C. Mozer, M.~Jordan, and T.~Petsche, editors, {\em NIPS 97},
  pages 1040--1046. MIT Press, Cambridge, MA, 1997.

\bibitem{scholkopf98}
Bernhard Sch\"{o}lkopf, Chris Burges, and Alex Smola.
\newblock Introduction to support vector learning.
\newblock In Bernhard Sch\"{o}lkopf, Chris Burges, and Alex Smola, editors,
  {\em Advances in Kernel Methods -- Support Vector Learning}, pages 1--22. MIT
  Press, 1998.

\bibitem{SteilHeidemannJockuschRaeJungclausRitter2001-GAF}
J.~J. Steil, G.~Heidemann, J.~Jockusch, R.Rae, N.~Jungclaus, and H.~Ritter.
\newblock Guiding attention for grasping tasks by gestural instruction: The
  {GRAVIS}-robot architecture.
\newblock In {\em Proc. IROS 2001}, pages 1570--1577. IEEE, 2001.

\bibitem{SuBaBook98}
Richard~S. Sutton and Andrew~G. Barto.
\newblock {\em Reinforcement Learning: An Introduction}.
\newblock MIT Press, Cambridge, MA, 1998.

\bibitem{TheSmi94}
Esther Thelen and Linda~B. Smith.
\newblock {\em A Dynamic Systems Approach to the Development of Cognition and
  Action}.
\newblock MIT Press, Cambridge, 1994.

\bibitem{EssenDeYoe95}
David~C. van Essen and Edgar~A. DeYoe.
\newblock Concurrent processing in primate visual cortex.
\newblock In Michael~S. Gazzaniga, editor, {\em The Cognitive Neurosciences},
  pages 383--400. MIT Press, 1995.

\bibitem{Sethu2001}
S.~Vijayakumar, J.~Conradt, T.~Shibata, and S.~Schaal.
\newblock Overt visual attention for a humanoid robot.
\newblock In {\em Proc IROS}, pages 2332--2337, 2001.

\bibitem{WalRit96}
J.~Walter and H.~Ritter.
\newblock Rapid learning with parametrized self-organizing maps.
\newblock {\em Neurocomputing}, 12:131--153, 1996.

\bibitem{HandAndBrain96}
Alan~M. Wing, Patrick Haggard, and J.~Randall Flanagan, editors.
\newblock {\em Hand and Brain}.
\newblock Academic Press, San Diego, 1996.

\bibitem{NI00}
M.~P. Young, editor.
\newblock {\em Special issue: Brain-structure-function relationships: advances
  from neuroinformatics}, volume 355 of {\em B}.
\newblock 2000.

\bibitem{Young93}
Malcolm~P. Young and Jack~W. Scannel.
\newblock Analysis and modelling of the organization of the mammalian cerebral
  cortex.
\newblock In Hans~G. Othmer, Philip~K. Maini, and James~D. Murray, editors,
  {\em Experimental and Theoretical Advances in Biological Pattern Formation},
  volume 259 of {\em NATO ASI Series A}, pages 369--384. Plenum Press, New
  York, 1993.

\end{thebibliography}

\end{document}